\documentclass[10pt, lettersize, journal, onecolumn]{IEEEtran/IEEEtran}
\usepackage{amsfonts}
\usepackage{algorithmic}
\usepackage{algorithm}
\usepackage{array}
\usepackage[caption=false, font=normalsize, labelfont=sf, textfont=sf]{subfig}
\usepackage{textcomp}
\usepackage{stfloats}
\UseRawInputEncoding
\usepackage{url}
\usepackage{verbatim}
\usepackage{graphicx}
\usepackage{cite} 
\usepackage{xcolor} 
\usepackage[hidelinks]{hyperref}
\usepackage{makecell} 
\usepackage{booktabs} 
\usepackage{tabularx} 
\usepackage{geometry} 
\usepackage{amsmath}
\usepackage[acronym,nogroupskip,nopostdot]{glossaries}
\makenoidxglossaries
\usepackage{xspace}
\usepackage{mfirstuc}
\newcommand{\pccpConcept}{infrastructure\xspace}

\newcommand{\approve}{APPROVE\xspace}
\newcommand{\condApprvl}{\mbox{CONDITIONAL} \mbox{APPROVAL}\xspace}
\newcommand{\clinRev}{CLINICAL \mbox{REVIEW}\xspace}
\newcommand{\reject}{REJECT\xspace}
\newcommand{\alarm}{ALARM\xspace}

\newcommand{\fixedPerfRef}{fixed performance reference\xspace}
\newcommand{\fixedPerfRefLong}{performance of the first fully approved model on the Golden dataset\xspace} 
\newcommand{\fixedPerfRefTol}{fixed performance reference tolerance\xspace}
\newcommand{\PgRelDef}{released performance reference\xspace}
\newcommand{\PgdRel}{fully approved released model performance on Golden/Drifting set\xspace}
\newcommand{\PgRelDefLong}{most recently fully approved released model performance on Golden set\xspace}
\newcommand{\PdRelDef}{most recently fully approved released model performance on Drifting set\xspace} 
\newcommand{\PPMSSaftyFloor}{\gls{pms} safety performance floor\xspace}
\newcommand{\PfailSaftyFloor}{deployment safety performance floor\xspace}
\newcommand{\PfailPMS}{deployment/\gls{pms} safety performance floor below which \reject/\alarm is triggered\xspace}
\newcommand{\PPMSDefLong}{\gls{pms} safety performance floor below which \alarm is triggered\xspace}
\newcommand{\PfailDefLong}{deployment safety performance floor below which \reject is triggered\xspace}
\newcommand{\PgoldDrift}{candidate model performance on Golden/Drifting dataset at iteration $k$\xspace}
\newcommand{\PgCurDef}{candidate model performance on Golden dataset at iteration $k$\xspace}
\newcommand{\tolDef}{acceptable deviation from target performance (non-inferiority margin)\xspace}
\newcommand{\driftScoreDef}{drift score\xspace}
\newcommand{\BonferroniDriftScoreDef}{Bonferroni-corrected \gls{ks} drift score\xspace} 
\newcommand{\PpmsPfailDistinction}{Production implementations may set $\PPMS > \Pfail$ to provide an early field warning before the deployment rejection threshold is reached; the two thresholds serve different governance functions and are evaluated on different models against different datasets.\xspace}
\newcommand{\benchmarkModelPerf}{performance of the model trained on the full cohort on the Golden set\xspace}
\newcommand{\benchmarkModel}{model trained on the full cohort\xspace}
\newcommand{\thrSource}{published clinical performance benchmarks and formal risk assessment per ISO 14971:2019\xspace}
\newcommand{\thrDisclaimer}{All thresholds are illustrative; production values require ISO 14971:2019 hazard analysis with documented traceability and/or alignment with published clinical performance benchmarks (see Supplementary S2.3).\xspace}
\newcommand{\isoRisk}{ISO 14971:2019\xspace}
\newcommand{\driftScoreMinorMajor}{range for minor/major $\driftScore$\xspace}

\newcommand{\trainData}{\mathcal{D}_{T,k}} 
\newcommand{\goldData}{\mathcal{D}_{G}} 
\newcommand{\driftData}{\mathcal{D}_{D,k}} 
\newcommand{\tauval}{{\tau}} 
\newcommand{\CImarging}{{\delta_m}} 
\newcommand{\cSub}[1]{C_{#1}} 
\newcommand{\Pfail}{{P^{\text{fail}}}} 
\newcommand{\PPMS}{{P^{\text{PMS}}}} 
\newcommand{\driftScore}{\mathcal{S}_{D}} 
\newcommand{\driftScoreMinor}{{\mathcal{S}_{D}^{\text{minor}}}} 
\newcommand{\driftScoreMajor}{{\mathcal{S}_{D}^{\text{major}}}} 
\newcommand{\taiScore}{\mathcal{S}_{T}} 
\newcommand{\taiScoreThr}{{\mathcal{S}_{T}^{\text{thr}}}} 
\newcommand{\Pref}{{P^{\text{ref}}}} 

\newacronym{aegis}{AEGIS}{AI/ML Evaluation and Governance Infrastructure for Safety}
\newacronym{capa}{CAPA}{Corrective and Preventive Action}
\newacronym{psur}{PSUR}{Periodic Safety Update Report}
\newacronym{pccp}{PCCP}{Predetermined Change Control Plan}
\newacronym{pms}{PMS}{Post-Market Surveillance}
\newacronym{eu}{EU}{European Union}
\newacronym{mdr}{MDR}{Medical Devices Regulation}
\newacronym{fda}{FDA}{Food and Drug Administration}
\newacronym{ai}{AI}{Artificial Intelligence}
\newacronym{darm}{DARM}{Dataset Assimilation and Retraining Module}
\newacronym{mmm}{MMM}{Model Monitoring Module}
\newacronym{mlcps}{MLcps}{ML Cumulative Performance Score}
\newacronym{roc-auc}{ROC-AUC}{Area Under the ROC Curve}
\newacronym{ks}{KS}{Kolmogorov-Smirnov}
\newacronym{shap}{SHAP}{SHapley Additive exPlanations}
\newacronym{kpi}{KPI}{Key Performance Indicator}
\newacronym{cdm}{CDM}{Conditional Decision Module}
\newacronym{ci}{CI}{Confidence Interval}
\newacronym{pmcf}{PMCF}{Post-Market Clinical Follow-up}
\newacronym{ml}{ML}{Machine Learning} 
\newacronym{pr-auc}{PR-AUC}{Area Under the PR Curve}
\newacronym{ppv}{PPV}{Positive Predictive Value}
\newacronym{npv}{NPV}{Negative Predictive Value}
\newacronym{ehr}{EHR}{Electronic Health Record}
\newacronym{rf}{RF}{Random Forest} 
\newacronym{oob}{OOB}{Out-Of-Bag}
\newacronym{mpmri}{mpMRI}{multi-parametric MRI}
\newacronym{dsc}{DSC}{Dice Similarity Coefficient} 
\newacronym{ca}{CA}{Conformity Assessment} 
\newacronym{tai}{TAI}{Trustworthy AI}

\newacronym{mlmd}{MLMD}{ML/AI-enabled Medical Device} 
\newacronym{gmlp}{GMLP}{Good Machine Learning Practice} 
\newacronym{mdai}{MDAI}{Medical Device Artificial Intelligence} 
\newacronym{tplc}{TPLC}{Total Product Life Cycle}  
\newacronym{mhra}{MHRA}{Medicines and Healthcare products Regulatory Agency}  
\newacronym{mdcg}{MDCG}{Medical Device Coordination Group} 
\newacronym{faq}{FAQ}{Frequently Asked Questions}  
\newacronym{ivdr}{IVDR}{In vitro Diagnostic Medical Devices Regulation} 
\newacronym{imdrf}{IMDRF}{International Medical Device Regulators Forum}
\newacronym{aib}{AIB}{Artificial Intelligence Board}
\newacronym{sota}{SOTA}{State-Of-The-Art}
\newacronym{se}{SE}{Substantial Equivalence}
\newacronym{cdss}{CDSS}{Clinical Decision Support Systems}
\newacronym{hit}{HIT}{Health Information Technology}
\newacronym{samd}{SaMD}{Software as a Medical Device}
\newacronym{pma}{PMA}{Pre-Market Approval}
\newacronym{gli}{GLI}{Glioma}
\newacronym{t1n}{T1n}{Native T1-weighted}
\newacronym{t1c}{T1c}{Contrast-enhanced T1-weighted}
\newacronym{t2w}{T2w}{T2-Weighted}
\newacronym{t2f}{T2f}{T2-weighted Fluid-attenuated inversion recovery}
\newacronym{et}{ET}{Enhancing Tumor}
\newacronym{netc}{NETC}{Non-Enhancing Tumor Core}
\newacronym{ed}{ED}{Edema}
\newacronym{mri}{MRI}{Magnetic Resonance Imaging}
\newacronym{rc}{RC}{Resection Cavity}
\newacronym{tc}{TC}{Tumor Core} 
\newacronym{cinc}{CinC}{Computing in Cardiology} 
\newacronym{gbm}{GBM}{Gradient Boosting Machine}
\newacronym{icu}{ICU}{Intensive Care Unit}
\newacronym{fnr}{FNR}{False Negative Rate}
\newacronym{fpr}{FPR}{False Positive Rate}
\newacronym{tp}{TP}{True Positive}
\newacronym{tn}{TN}{True Negative}
\newacronym{fp}{FP}{False Positive}
\newacronym{fn}{FN}{False Negative}
\newacronym{miccai}{MICCAI}{Medical Image Computing and Computer Assisted Intervention}
\newacronym{brats}{BraTS}{Brain Tumor Segmentation}
\newacronym{abcds}{ABCDS}{Algorithm-Based Clinical Decision Support}
\newacronym{irb}{IRB}{Institutional Review Board}
\newcommand{\Pprev}{{P_{G, k-1}}} 
\newcommand{\Pcur}{{P_{G, k}}} 
\newcommand{\PDcur}{{P_{D, k}}} 
\newcommand{\PDprev}{{P_{D, k-1}}} 
\newcommand{\PgRel}{{P^{\text{rel}}_{G}}} 
\newcommand{\PdRel}{{P^{\text{rel}}_{D}}} 
\newcommand{\dSub}[1]{D_{#1}} 
\newcommand{\cNR}{{C^{\text{NR}}}} 
\newcommand{\cReg}{{C^{\text{R}}}} 
\newcommand{\RaccG}{{R_{G}}} 
\newcommand{\RaccD}{{R_{D}}} 
\newcommand{\trainDataNext}{\mathcal{D}_{T,k+1}} 
\newcolumntype{Y}{>{\raggedright\arraybackslash}m{\hsize}}
\newcommand{\glsposs}[1]{\gls{#1}'s} 
\newcommand{\condFunc}{{f}} 
\newcommand{\prio}[1]{{P{#1}}} 
\begin{document}
\title{
AEGIS: An Operational Infrastructure for Post-Market Governance of Adaptive Medical AI Under US and EU Regulatory Frameworks
}
\author{
\IEEEauthorblockN{Fardin Afdideh\IEEEauthorrefmark{1}, Mehdi Astaraki\IEEEauthorrefmark{1}\IEEEauthorrefmark{2}\IEEEauthorrefmark{3}, Fernando Seoane\IEEEauthorrefmark{1}\IEEEauthorrefmark{4}\IEEEauthorrefmark{5}\IEEEauthorrefmark{6}, and Farhad Abtahi\IEEEauthorrefmark{1}\IEEEauthorrefmark{4}\IEEEauthorrefmark{7}} \\
\IEEEauthorblockA{\IEEEauthorrefmark{1}Department of Clinical Science, Intervention and Technology, Karolinska Institutet, 17177 Stockholm, Sweden}\\
\IEEEauthorblockA{\IEEEauthorrefmark{2}Department of Medical Radiation Physics, Stockholm University, Solna, Sweden}\\
\IEEEauthorblockA{\IEEEauthorrefmark{3}Department of Oncology-Pathology, Karolinska Institutet, Solna, Sweden}\\
\IEEEauthorblockA{\IEEEauthorrefmark{4}Department of Clinical Physiology, Karolinska University Hospital, 17176 Stockholm, Sweden}\\
\IEEEauthorblockA{\IEEEauthorrefmark{5}Department of Textile Technology, University of Bor\aa{}s, Bor\aa{}s 50190, Sweden}\\
\IEEEauthorblockA{\IEEEauthorrefmark{6}Department of Medical Technologies, Karolinska University Hospital, 141 57 Huddinge, Sweden}\\
\IEEEauthorblockA{\IEEEauthorrefmark{7}Department of Biomedical Engineering and Health System, KTH Royal Institute of Technology, 14157 Huddinge, Sweden}
\thanks{Corresponding author: Farhad Abtahi (farhad.abtahi@ki.se).}
}
\maketitle
\begin{abstract}
Machine learning and artificial intelligence systems deployed in medical devices require governance frameworks that ensure safety while enabling continuous improvement. 
Regulatory bodies, including the U.S. \gls{fda} and the European Union, have introduced mechanisms to manage iterative model updates without repeated submissions. The \gls{pccp} and \gls{pms} requirements exemplify this approach. This paper presents \gls{aegis}, a governance \pccpConcept applicable to any healthcare \gls{ai} system. \gls{aegis} comprises three modules: the dataset assimilation and retraining module, model monitoring module, and conditional decision module. 
Together, these operationalize \gls{fda} \gls{pccp} requirements and \gls{eu} \gls{ai} Act Article 43(4) provisions for predetermined changes. 
\gls{aegis} specifies \textit{what} must be monitored and \textit{how} decisions must be made, while clinical context determines thresholds, metrics, and architectures. 
We implement a four-category deployment decision taxonomy (\approve, \condApprvl, \clinRev, \reject) combined with an independent \gls{pms} \alarm signal, and a \fixedPerfRef comparison mechanism for regulatory \gls{se}. The composite output enables detection of the critical governance state in which no deployable new model is available and the currently released model is simultaneously assessed as at risk. 
To demonstrate generalizability, we applied \gls{aegis} to two heterogeneous domains: sepsis prediction from electronic health records and brain tumor segmentation from medical imaging. 
Identical governance architecture accommodated different data modalities through configuration alone. 
Demonstrated across the PhysioNet Sepsis Challenge (20,336 patients from Hospital A; 4,067 reserved as the Golden evaluation set) across 11 iterations yielded 8 \approve decisions (iterations 0--5 and 8--9), 1 \condApprvl from minor cross-site drift (iteration 6), 1 \clinRev from regression from \fixedPerfRef  (iteration 7), and 1 \reject when sensitivity (0.428) fell below the 0.65 threshold, exercising all four deployment decision categories. \alarm signals were co-issued at iterations 8 (\approve + \alarm: new model deployable but released model at risk from major distributional shift) and 10 (\reject + \alarm: no deployable model and released model simultaneously failing, representing the critical safety escalation state).
Across iterative deployment simulations, \gls{aegis} consistently detected distributional drift before observable performance degradation and automatically enforced safety-critical thresholds, co-issuing \gls{pms} \alarm signals when the released model was assessed as at risk and triggering \reject decisions when predefined clinical performance floors were violated. 
Ultimately, these results demonstrate that \gls{aegis} provides a practical operational layer that translates high-level \gls{fda} and \gls{eu} regulatory change-control concepts into executable governance procedures. By successfully preventing silent degradation across both structured 2D and unstructured 3D data without requiring architectural modifications, \gls{aegis} supports safe, continuous learning for adaptive medical \gls{ai} systems across diverse clinical applications.
\end{abstract} 
\section{Introduction} 
\noindent
\gls{ml} models in healthcare differ fundamentally from traditional medical devices: they learn and adapt over time. This creates a regulatory challenge. When an algorithm's performance drifts, current frameworks provide no systematic response mechanism. The \gls{fda}'s adverse event database cannot capture gradual degradation because it does not constitute a reportable event \cite{babic_general_2025}.

Over 1,400 AI-enabled medical devices have received \gls{fda} clearance or authorization since the first approval in 1995 \cite{health_artificial_2025}. These systems increasingly support clinical decision-making in diverse settings, from imaging diagnostics to acute risk estimation. Yet the regulatory infrastructure designed for static devices struggles to accommodate algorithms that improve through real-world use.

The \gls{fda} recognized this mismatch and initiated a policy shift. 
Beginning with a 2019 discussion paper \cite{health_artificial_2025-3} and culminating in landmark guidance in August 2025 \cite{health_marketing_2025}, the agency established the \gls{pccp} framework. 
Under a \gls{pccp}, manufacturers prespecify acceptable modifications and the criteria for implementing them. Once approved, changes within these bounds require no additional submissions \cite{health_marketing_2025}. This applies across major pathways: 510(k) \cite{health_premarket_2024}, De Novo \cite{health_novo_2025}, and Pre-Market Approval \cite{health_premarket_2023}.

The \gls{eu} \gls{ai} Act contains parallel provisions. Article 43(4) permits ``predetermined changes'' that avoid triggering new \glspl{ca} when documented in technical documentation \cite{noauthor_regulation_2024}. However, ambiguity persists around what constitutes a ``substantial modification'' requiring reassessment. The Medical Device Coordination Group and \gls{ai} Board have begun issuing joint guidance to clarify implementation \cite{noauthor_mdcg_nodate}. 

Both regulatory frameworks share a critical implementation gap: they specify \textit{what} manufacturers must do without defining \textit{how} to do it operationally. Four questions remain unanswered:
\begin{enumerate}
    \item What specific metrics should trigger a model update or rollback?
    \item What quantitative thresholds define acceptable versus unacceptable performance?
    \item How should drift detection be operationalized with statistical rigor?
    \item What audit trail structure ensures regulatory defensibility?
\end{enumerate}
Translating these regulatory concepts into executable governance procedures remains an unsolved challenge.

Babic et al.\ \cite{babic_general_2025} quantified this gap through systematic analysis of the \gls{fda}'s MAUDE database. Their review of 943 adverse events across 823 AI/ML devices (2010--2023) found the surveillance infrastructure ``insufficient for properly assessing the safety and effectiveness of AI/ML devices.'' 
Critically, concept drift and covariate shift cannot be reported because they ``do not constitute an adverse event.'' Aggregate performance deterioration remains invisible to case-based reporting.

While regulatory guidance establishes the need for continuous \gls{pms} of adaptive medical \gls{ai} \cite{health_good_2025}, the operational decision logic required for change control remains unspecified, a gap echoed in recent governance frameworks \cite{li_prioritizing_2026}.
This paper addresses these gaps by presenting \gls{aegis}, a technical \pccpConcept that operationalizes \gls{pccp} requirements through executable specifications. 
\gls{aegis} provides three core components: a \gls{darm} for data governance, a \gls{mmm} for continuous surveillance, and a \gls{cdm} for automated governance decisions.

\gls{aegis} parallels the \gls{fda}'s concept of \gls{se} used in 510(k) submissions \cite{health_premarket_2024}. 
In the 510(k) process, manufacturers demonstrate that a new device is as safe and effective as a predicate device. 
\gls{aegis} implements an analogous ``\fixedPerfRef comparison mechanism'' for model updates. New iterations must perform within defined bounds of the \fixedPerfRef metrics or demonstrate improvement.

\subsection{Contributions}
\noindent
This paper makes the following contributions:
\begin{enumerate}
 \item A domain-agnostic governance architecture comprising three modules (\gls{darm}, \gls{mmm}, \gls{cdm}) that operationalize \gls{fda} \gls{pccp} requirements and \gls{eu} \gls{ai} Act Article 43(4) provisions. The \pccpConcept specifies monitoring requirements and decision logic; clinical context determines thresholds and metrics through configuration.
 
 \item A four-category deployment decision taxonomy (\approve, \condApprvl, \clinRev, \reject) combined with an independent \gls{pms} \alarm signal, aligned with US and \gls{eu} regulatory guidance. 
 The deployment decision taxonomy governs change-control determinations documentable in \gls{pccp} submissions; the \alarm signal addresses \gls{pms} obligations under \gls{eu} \gls{mdr} Article 83 and \gls{fda} \gls{pms} expectations. 
 The composite output captures the critical \reject + \alarm state, where no deployable new model is available and the released model is simultaneously assessed as at risk in the field, a governance condition unrepresentable in prior frameworks. 
 
 \item A \fixedPerfRef comparison mechanism implementing regulatory \gls{se}. New model iterations must satisfy acceptance criteria relative to \fixedPerfRef metrics. The \pccpConcept supports both deterministic thresholds and \gls{ci}-based equivalence testing.
 
 \item Integration of the \gls{mlcps} \cite{akshay_mlcps_2023} into governance workflows. This composite metric enables domain-specific weighting for unified performance assessment.
 
 \item Demonstration across two heterogeneous domains: sepsis prediction (tabular time-series, \gls{rf}) and brain tumor segmentation (volumetric imaging, convolutional networks). The same architecture accommodates both through configuration alone.
\end{enumerate}

The clinical applications serve as demonstration vehicles. The underlying models are intentionally conventional; the contribution lies in the governance infrastructure.

The remainder of this paper proceeds as follows. Section II describes the \gls{aegis} architecture, including the three core modules and the decision taxonomy. Section III presents validation results from both clinical domains. Section IV discusses regulatory implications, limitations, and future directions.

\section{Materials and Methods} 
\noindent

This section presents \gls{aegis} in two parts: (1) \gls{aegis} Architecture, describing domain-agnostic governance components, and (2) \gls{aegis} Instantiation, demonstrating configuration for specific clinical domains.

\subsection{Regulatory Context and Design Rationale}
\noindent
Two regulatory mandates drive \gls{aegis} design: auditable change control and rigorous \gls{pms} for continuously learning models.

The \gls{pccp} requires manufacturers to define, a priori, the criteria under which models may be modified and the acceptable limits of performance change. This necessitates a formal gate that objectively determines whether an update requires new submission or falls within pre-authorized scope \cite{health_predetermined_2025}. 

The \gls{eu} \gls{mdr} 2017/745 demands comprehensive \gls{pms} and clinical follow-up. For adaptive models, this means continuously monitoring real-world performance \cite{noauthor_regulation_2017}, demonstrating that new data does not introduce unacceptable bias \cite{health_good_2025}, and ensuring safety claims remain valid across releases.

The \gls{eu} \gls{ai} Act emphasizes quality management, data governance, and robustness for high-risk applications. It requires continuous monitoring and documented control of changes to system logic or data distributions \cite{noauthor_regulation_2024}.

These frameworks differ in emphasis. The \gls{pccp} focuses on pre-defining acceptable changes; the \gls{mdr} emphasizes ongoing risk assessment; the \gls{ai} Act mandates extensive documentation. Table \ref{tab:regulatory_philosophy} summarizes the philosophical differences between \gls{fda} and \gls{eu} approaches.

\begin{table*}[t]
    \centering
    \caption{Regulatory Philosophy Comparison: \gls{fda} vs.\ \gls{eu} \gls{mdr}}
    \label{tab:regulatory_philosophy}
    \begin{tabularx}{\textwidth}{m{3cm} X X}
        \toprule
        \multicolumn{1}{c}{\textbf{Dimension}} & \multicolumn{1}{c}{\textbf{\gls{fda} (510(k)/\gls{pccp})}} & \multicolumn{1}{c}{\textbf{\gls{eu} \gls{mdr}}} \\
        \midrule
        Core Philosophy & \gls{se} & Continuous risk management \\
        \midrule
        Foundational Question & ``Is this device as safe and effective as something already proven?'' & ``Does the ongoing risk-benefit analysis remain acceptable?'' \\
        \midrule
        Evidence Paradigm & Comparative (new vs.\ predicate) & Cumulative (ongoing real-world evidence) \\
        \midrule
        Clinical Data Focus & Premarket; 510(k) may rely on bench testing & Continuous clinical evaluation (\gls{pmcf}) \\
        \midrule
        Change Control Logic & Modifications within \gls{pccp} bounds maintain equivalence & Changes assessed against evolving risk profile \\
        \midrule
        Methodological Backbone & Predicate device comparison & \isoRisk risk management \\
        \bottomrule
    \end{tabularx}
\end{table*}

The \gls{fda} grounds safety determinations in \gls{se}: devices gain clearance by demonstrating comparability to legally marketed predicates. The \gls{pccp} extends this philosophy to post-market modifications. In contrast, the \gls{mdr} anchors to continuous risk management per \isoRisk, asking whether ongoing risk-benefit analysis remains acceptable throughout the lifecycle.

\gls{aegis} addresses both paradigms simultaneously. 
The \fixedPerfRef comparison mechanism is designed to support \gls{fda} equivalence requirements. The \gls{mmm}'s continuous monitoring is structured to align with \gls{mdr} surveillance mandates. 
The four-category deployment taxonomy combined with an independent \gls{pms} \alarm signal maps to both structures.

\subsection{\gls{aegis} Architecture}
\noindent
\gls{aegis} defines governance architecture through three modules. This architecture specifies \textit{what} must be monitored and \textit{how} decisions must be made. Domain-specific adaptation requires only threshold calibration and metric weighting, not architectural changes.

\gls{aegis} maps directly to the \gls{fda} \gls{pccp} components as defined in the August 2025 final guidance \cite{health_marketing_2025, health_predetermined_2025}. 
The \textbf{Description of Modifications} is addressed through \gls{darm} protocols and \gls{cdm} decision categories. The \textbf{Modification Protocol} is realized through \gls{mmm} metric procedures. The \textbf{Impact Assessment} is operationalized through \fixedPerfRef comparison.

\subsubsection{Dataset Assimilation and Retraining Module (DARM)} 
\noindent
In geosciences, data assimilation integrates new observations with numerical models to refine system estimates \cite{kalnay_ncepncar_1996}. We adapt this concept for ML governance. The \gls{darm} manages controlled ingestion and structured integration of new observations, assigning definitive status to every data point.
Before these newly ingested observations can be assigned to their respective dataset splits, they must first undergo rigorous quality control and high-confidence annotation to establish an accurate and auditable clinical ground truth.
\paragraph{Label Governance and Expert Review}
\noindent
Label provenance and adjudication are governed within the quality management system. Incoming cases are labelled using pre-specified criteria. A subset undergoes expert adjudication with documented disagreement resolution. Cases flagged as uncertain or out-of-distribution are excluded until resolved. Each batch is versioned with provenance metadata (source, collection window, labelling method, reviewer identifiers, timestamps).

This mechanism ensures dataset integrity (strict separation between training, validation, and evaluation data) and transparency (auditable logs of collection timing and allocation). Three dataset splits are maintained at each iteration:

\paragraph{Training Dataset} 
\noindent
The Training dataset updates model parameters at each iteration. This dynamic set comprises examples reviewed by expert clinicians with high label confidence. At iteration $k$, the model trains exclusively on $\trainData$.

\paragraph{Golden Dataset} 
\noindent
The Golden dataset ($\goldData$) is a static, held-out benchmark for evaluating generalization across all iterations. It is never used for training, validation, or tuning. This guarantees a consistent, unbiased comparison across iterations.

\paragraph{Drifting Dataset} 
\noindent
The Drifting dataset ($\driftData$) is dynamic and serves two functions at each iteration $k$:

\begin{itemize}
    \item Drift Testing: $\driftData$ tests the current model internally. Performance on this set is monitored for distribution changes or feature decay.
    \item Subsequent Training Pool: Upon successful testing and expert review, $\driftData$ contents integrate into $\trainData$ for iteration $k+1$, enabling continuous learning.
\end{itemize}

The \gls{darm} addresses \gls{eu} \gls{ai} Act data governance requirements, \gls{mdr} evidence needs, and \gls{pccp} control over retraining data. By assigning status to every observation, it provides auditable logs preventing evaluation set contamination.

\begin{figure}[h!]
 \centering
 \includegraphics[width=1\textwidth]{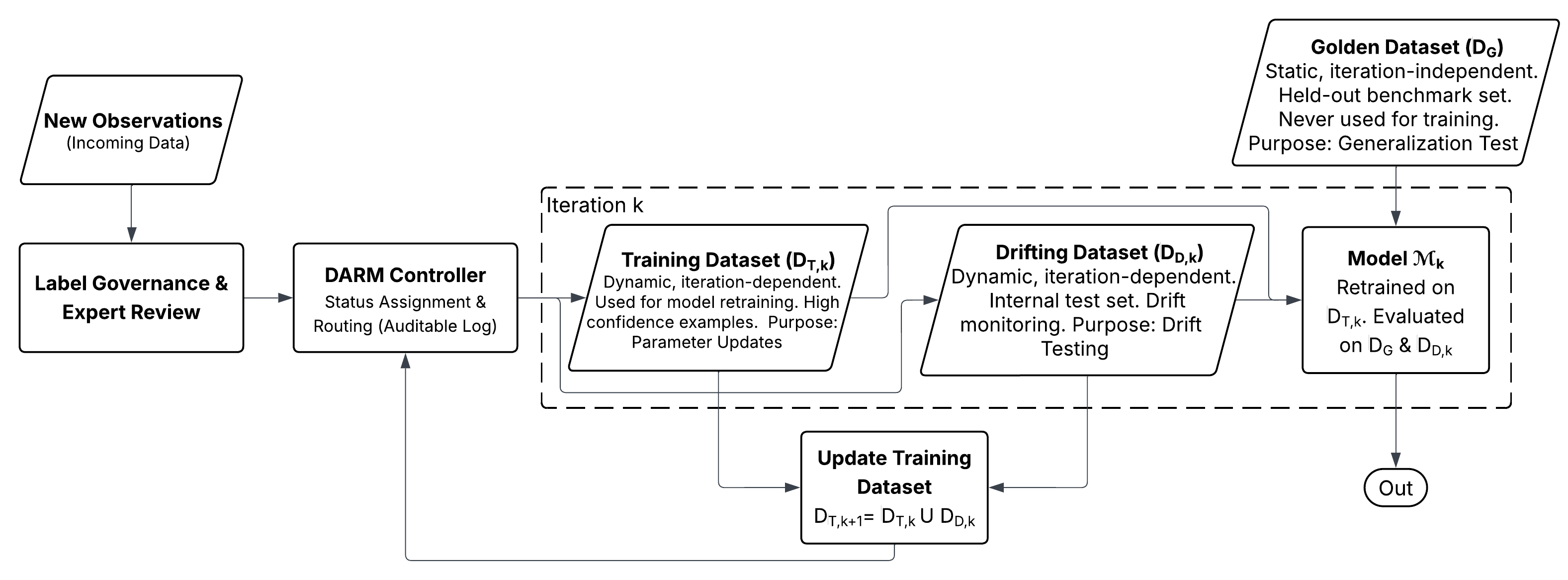}
 \caption{\gls{darm} architecture. The module manages three datasets: Training ($\trainData$) for parameter updates, Golden ($\goldData$) as static benchmark, and Drifting ($\driftData$) for drift monitoring. 
 }
 \label{fig:dam_block_diagram}
\end{figure}

Figure \ref{fig:dam_block_diagram} illustrates the \gls{darm} architecture, showing dataset interactions and the feedback loop where $\driftData$ integrates into $\trainDataNext$.

\subsubsection{Model Monitoring Module (MMM)} 
\noindent
The \gls{mmm} calculates metrics on dataset splits and maintains historical records of \glspl{kpi}. Table \ref{tab:parameters} summarizes required parameter categories.

\begin{table}[h]
    \small
 \centering
 \caption{Parameters}
 \label{tab:parameters}
 \begin{tabular}{ll}
 \toprule
  \multicolumn{1}{c}{\textbf{Parameter Categories}} & \multicolumn{1}{c}{\textbf{Description}} \\
 \midrule
 Tolerance & Acceptable deviation from target value\\
 \midrule
  Range & Minimum and maximum bounds for acceptable values\\
 \midrule 
 Reference & Reference values against which comparisons are made \\ 
 \midrule
 Distribution Shift Metrics & KL divergence, JS distance, mean drift\\
 \midrule
 \gls{tai} Metrics & Demographic parity, equalized odds metrics\\
 \midrule
  Model Performance Metrics & Accuracy, sensitivity, specificity, AUC, F1-score\\
 \bottomrule
 \end{tabular}
 \begin{flushleft}  
     \footnotesize{Note: The constant values (tolerance, range, and reference) are derived during \gls{pccp} preparation from \thrSource.}
\end{flushleft} 
\end{table}

\begin{figure}[h!]
 \centering
 \includegraphics[width=1\textwidth]{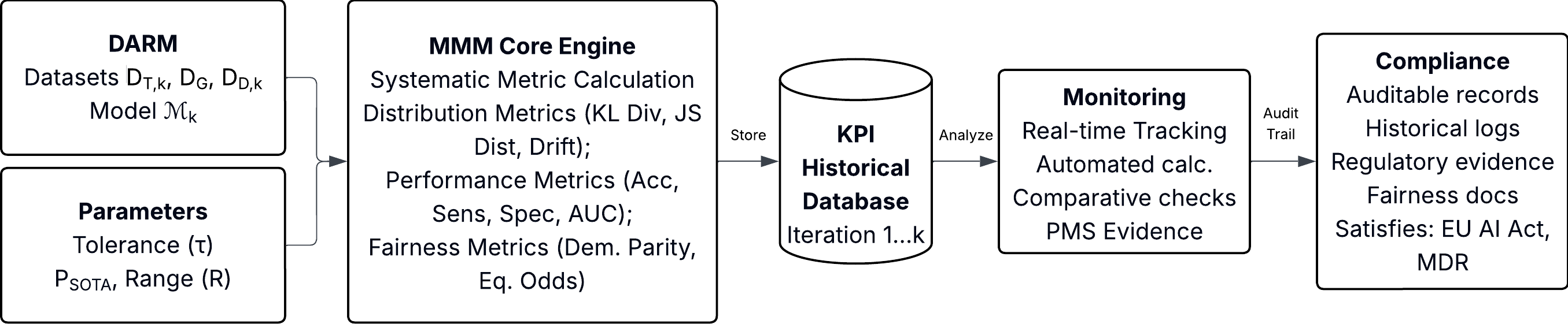}
 \caption{\gls{mmm} architecture showing metric calculation across performance, fairness, and distribution categories, historical \gls{kpi} maintenance, and outputs for monitoring and compliance.}
 \label{fig:mmm_block_diagram}
\end{figure}

The \gls{mmm} provides unbiased evidence for continuous monitoring and drift detection by calculating performance metrics on both static and dynamic sets from the \gls{darm}. Figure \ref{fig:mmm_block_diagram} illustrates the architecture.

\subsubsection{Conditional Decision Module (CDM)} 
\noindent
The \gls{cdm} produces a composite output at each iteration: (1) a deployment decision (\approve, \condApprvl, \clinRev, \reject) governing the \textit{new candidate model}, and (2) a \gls{pms} \alarm signal assessing the continued safety of the \textit{currently released model already in use}. These two outputs are evaluated independently against separate metric sets and may be combined, which is why \approve + \alarm is not contradictory: it means the new model is ready to deploy while the old model is simultaneously showing signs of risk in the field.

To formalize the decision logic, we first define a set of evaluation conditions. 
Let $\cSub{i}$ denote condition $i$, for $i = 1,\ldots,N$, typically comparing a performance metric with a predefined threshold. 
The decision process can then be expressed using a ternary operator defined in Eq.~\eqref{eq:ternary}:
\begin{equation}
\label{eq:ternary}
    \dSub{} = \condFunc(\cSub{1}, \cSub{2}, \dots, \cSub{N}) \text{ } ? \text{ } \dSub{1} \text{ } : \text{ } \dSub{2},
\end{equation}
where $\condFunc$ is an aggregated Boolean function. If $\condFunc$ evaluates to \textit{True}, decision $\dSub{1}$ is selected; otherwise, $\dSub{2}$.
The formulation presented in Eq.~\eqref{eq:ternary} supports arbitrary logical combinations. 
For example, for $\dSub{1} = \text{``\approve''}$ and $\dSub{2} = \text{``OTHER''}$, one may define $\condFunc$ as $(\cSub{1} \land \cSub{2} \land \cSub{3})$, where $\cSub{1} = (\Pcur \geq \Pref - \tauval)$ tests proximity to optimal performance, $\cSub{2} = (\Pcur \geq \Pprev)$ tests non-regression, and $\cSub{3} = (\Pcur \in \RaccG) \land (\PDcur \in \RaccD)$ tests cross-split stability. 
Complete specifications, including decision logic pseudo-code and threshold configuration templates, appear in Supplementary Materials S2.6 and S3.3. 

Table \ref{tab:decision_variants} shows example decision configurations for the four deployment categories and the \gls{pms} \alarm signal.
Each row in Table \ref{tab:decision_variants} expresses the aggregated Boolean condition that characterizes one decision category or signal.

Escalation decisions (from \reject\ to \condApprvl) use failure-trigger conditions that evaluate to \textit{True} when a failure occurs; the natural aggregation operator is therefore logical disjunction ($\lor$), ensuring a response is triggered by any single failing condition. For an \approve\ decision, by contrast, conditions evaluate to \textit{True} when no failure is detected (i.e., pass-trigger conditions), making logical conjunction ($\land$) the appropriate operator: all checks must pass simultaneously to trigger approval.

When conditions for multiple categories are satisfied simultaneously, the priority-ordered hierarchy in Table \ref{tab:decision_hierarchy} determines which decision is returned. 
The two tables are complementary: Table \ref{tab:decision_variants} defines the logical structure of each category; Table \ref{tab:decision_hierarchy} defines the resolution order across categories.
\begin{table*}[!ht] 
  \small
  \centering
  \caption{\gls{cdm} Decision Categories. 
  Deployment decisions: Four decision categories for a new-model deployment. \gls{pms} \alarm signal: \alarm for the currently released model. The two parts are evaluated independently and produce a composite output.}
  \label{tab:decision_variants}
  \begin{tabularx}{\textwidth}{m{0.14\linewidth} m{0.2\linewidth} X X}
    \toprule
    \multicolumn{1}{c}{\textbf{Decision}} & \multicolumn{1}{c}{\textbf{Question answered}} & \multicolumn{1}{c}{\textbf{Boolean Function \condFunc}} & \multicolumn{1}{c}{\textbf{Operator \& Rationale}} \\
    \midrule
    \multicolumn{4}{l}{\textbf{CHANGE-CONTROL DEPLOYMENT DECISIONS: new candidate model}} \\  
    \midrule
    \reject & 
    Is new model safe? & 
    $(\Pcur < \Pfail) \lor (\text{any safety metric} < \text{its floor})$ & 
    OR --- any single safety failure blocks deployment \\
    \midrule
    \clinRev & 
    Does new model need human validation? & 
    $  
    (\Pfail \leq \Pcur < \Pfail + \tauval) \lor (\tauval < \Pref - \Pcur)$ & 
    OR --- any metric in buffer zone, or any regression from $\Pref$ (\fixedPerfRef comparison fails). 
    \\
    \midrule
    \condApprvl & 
    Does the new model need enhanced monitoring? & 
    $
    (\driftScore \in \driftScoreMinor) \lor (\taiScore < \taiScoreThr)
    $
    & 
    OR --- minor drift or any TAI concern each independently warrant enhanced monitoring \\
    \midrule
    \approve & 
    Does new model pass all checks? & 
    $ 
    ((\Pref - \tauval) \leq \Pcur) \land (\Pprev \leq \Pcur) \land (\driftScore \leq \driftScoreMinor) \land (\taiScoreThr \leq \taiScore)
    $ &  
    AND --- all checks must pass; any single failure routes to a lower category \\
    \midrule
    \multicolumn{4}{l}{\textbf{\gls{pms} \alarm SIGNAL: currently released model}} \\
    \midrule
    \alarm & 
    Is the released model still safe in the field? & 
    $(\PdRel \leq \PPMS^*) \lor (\PdRel \leq \PgRel - \tauval) \lor (\driftScore \in \driftScoreMajor)$
    & 
    OR --- released model $\PPMS$ (\PPMSSaftyFloor) breached, or released model regressed from \PgRelDef, or major drift threatens released model validity. \\
    \bottomrule
  \end{tabularx}
    \begin{flushleft}
        \footnotesize{Notations: $\Pcur$ = \PgCurDef; $\Pfail/\PPMS$ = \PfailPMS; $\tauval$ = \tolDef; $\driftScore$ = \driftScoreDef; $\driftScoreMinor/\driftScoreMajor$ = \driftScoreMinorMajor; $\taiScore$ = \gls{tai} composite score; $\taiScoreThr$ = threshold for $\taiScore$; $\Pref$ = \fixedPerfRef; $\PgRel/\PdRel$ = \PgdRel.} \\    
\footnotesize{$^*$\PpmsPfailDistinction.}\\      
         \footnotesize{Note: \textit{\thrDisclaimer}}
    \end{flushleft}
\end{table*}

\begin{table*}[!ht]
    \small
    \centering
    \caption{Decision Hierarchy: Priority-Ordered Evaluation}
    \label{tab:decision_hierarchy}
    \begin{tabularx}{\textwidth}{c X X l}
        \toprule
        \multicolumn{1}{c}{\textbf{Priority}} & \multicolumn{1}{c}{\textbf{Check}} & \multicolumn{1}{c}{\textbf{Condition}} & \multicolumn{1}{c}{\textbf{Decision}} \\
        \midrule
        \multicolumn{4}{l}{\textbf{New Model Deployment Hierarchy (evaluate in priority order on $\goldData$ and $\driftData$)}} \\ 
        \midrule
        \prio{1} & \PfailSaftyFloor  &  $\Pcur < \Pfail$ & \reject \\
        \midrule
        \prio{2} & performance buffer / regression from \fixedPerfRef  & $(\Pcur \in \RaccG) \lor (\tauval < \Pref - \Pcur)$ & \clinRev \\
        \midrule
        \prio{3} & minor drift / \gls{tai} & $(\driftScore \in \driftScoreMinor) \lor (\taiScore < \taiScoreThr)$ & \condApprvl \\
        \midrule
        \prio{4} & all pass & all checks passed & \approve \\
        \midrule
        \multicolumn{4}{l}{\textbf{\gls{pms} \alarm Signal (evaluate in parallel using released model on $\driftData$)}} \\ 
        \midrule
        A1 & \PPMSSaftyFloor & $\PdRel < \PPMS$ & \alarm \\
        \midrule
        A2 & released model regression & $\PdRel \leq \PgRel - \tauval$ & \alarm \\
        \midrule 
        A3 & major distributional shift & $\driftScore \in \driftScoreMajor$ & \alarm \\
        \bottomrule
    \end{tabularx}
    \vspace{0.5em}
    \begin{flushleft}
        \footnotesize{Notations: $\driftData$ = Drifting dataset at iteration $k$; $\goldData$ = Golden dataset; $\PdRel$ = \PdRelDef; $\driftScore$ = \driftScoreDef; $\driftScoreMinor/\driftScoreMajor$ = \driftScoreMinorMajor; $\taiScore$ = \gls{tai} composite score; $\taiScoreThr$ = threshold for $\taiScore$.}\\
        \footnotesize{Note I: The deployment decision and \alarm signal run in parallel; the composite output is (deployment decision) or (deployment decision + \alarm) when any of A1--A3 fires.}\\
        \footnotesize{Note II: All threshold parameters ($\Pfail$, $\PPMS$, $\tauval$,
        $\driftScoreMinor$, $\driftScoreMajor$, $\taiScoreThr$) are instantiation-defined
        \gls{pccp} parameters specified during deployment configuration per \isoRisk
        risk assessment. See Table~\ref{tab:sepsis_param_mapping} (sepsis) and
        Table~\ref{tab:bt_param_mapping} (BraTS) for domain-specific value examples.}
    \end{flushleft}
\end{table*}

Condition weighting is managed through threshold adjustment. Critical conditions (e.g., minimum sensitivity) require conservative thresholds reflecting low failure tolerance. Threshold specification must trace to \isoRisk risk management principles \cite{noauthor_iso_nodate}.

\textbf{$\PgRel$ update rule.} The \PgRelDefLong ($\PgRel$ in Table~\ref{tab:decision_variants}) is updated exclusively on \approve decisions. 
During periods of \condApprvl or \clinRev, $\PgRel$ remains anchored to the most recent full \approve, regardless of which model is currently deployed. 
This ensures that the regression check from $\PgRel$ is always evaluated against a fully validated reference, not against a tentative deployment (\condApprvl or \clinRev).

\textbf{$\PgRel$ \textit{vs} $\PdRel$ distinction.} $\PgRel$ is the \PgRelDefLong, while $\PdRel$ is the \PdRelDef.
The \gls{pms} ALARM signal fires when the gap between these two quantities exceeds a tolerance $\tauval$.

\textbf{$\PPMS$ \textit{vs} $\Pfail$ distinction.} $\PPMS$ is the \PPMSDefLong, while $\Pfail$ is the \PfailDefLong.
$\Pfail$ governs the new candidate model evaluated on the Golden dataset $\goldData$; $\PPMS$ governs the currently fully approved released model evaluated on the incoming drifting batch $\driftData$. 
Both are pre-specified in the \gls{pccp}, but they address different models, different datasets, and different governance questions. 
Setting $\Pfail < \PPMS$ in production creates a tiered safety architecture in which field risk is flagged before a formal deployment rejection threshold is reached.

\textbf{Decision confidence.} We define confidence as the bootstrap-estimated probability that an update satisfies acceptance criteria. For iteration $k$, we bootstrap evaluation sets $B$ times, recompute metrics and conditions for each replicate, and define confidence as the fraction yielding the same outcome as the point estimate.

\begin{figure}[h!]
 \centering
 \includegraphics[width=1\textwidth]{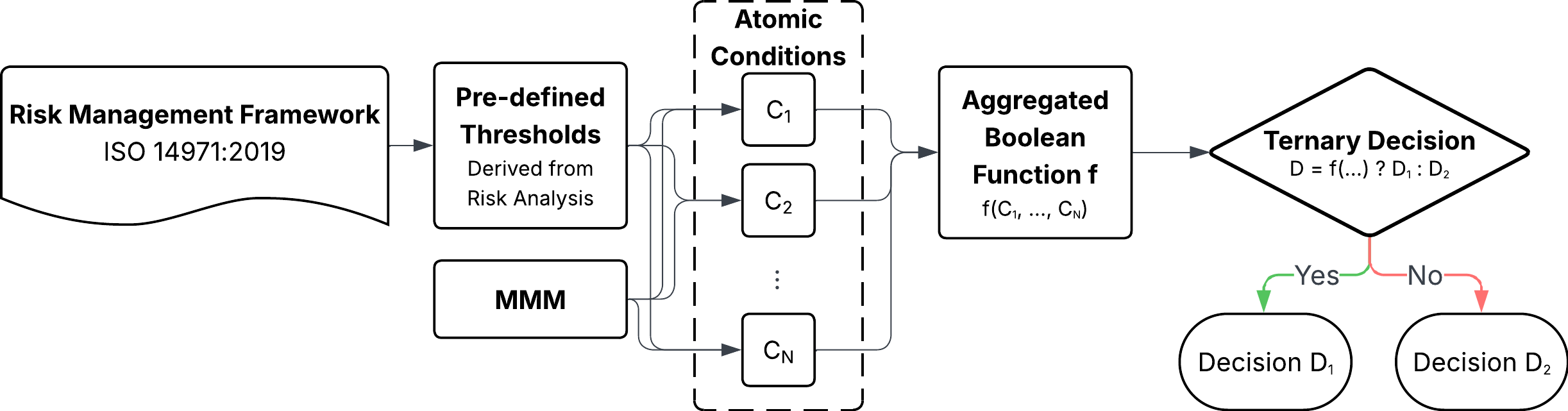}
 \caption{\gls{cdm} architecture showing flow from metrics and thresholds through atomic conditions and Boolean aggregation to decision outcomes.}
 \label{fig:cdm_block_diagram}
\end{figure}

Figure \ref{fig:cdm_block_diagram} illustrates the \gls{cdm} architecture.

\subsubsection{\capitalisewords{\fixedPerfRef} Comparison Mechanism} 
\noindent
Threshold values (e.g., $\Pfail$, $\tauval$, $\taiScoreThr$) are derived during \gls{pccp} preparation from \thrSource. 
After the first \approve decision, \gls{aegis} records the model's measured performance on the Golden dataset as the \fixedPerfRef ($\Pref$), which is an internally measured post-approval value; its numerical magnitude is informed by, but not identical to, any pre-specified external bound.
$\Pref$ serves as the invariant operational reference in subsequent \gls{cdm} evaluations using acceptance criteria described below.

In a \gls{pccp} context, $\Pref$ is derived from data available at the time of the device's initial regulatory-authorized release. Alternatively, it may be drawn from externally established benchmarks such as validated clinical performance standards or published reference values, provided these are available and consistent with the clinical task and dataset characteristics. In simulation settings where real deployment data is unavailable, $\Pref$ may be computed using the full available cohort, as is the case in the brain tumor segmentation instantiation presented in Section \ref{sec:BrainTumorSegInst}.

\subsubsection{Statistical Acceptance Criteria}
\noindent
\gls{aegis} supports two approaches for evaluating acceptable performance.

\paragraph{Deterministic threshold comparison} Point estimates are compared directly against thresholds. For each metric $m \in \{\text{sensitivity}, \text{specificity}, \text{\gls{roc-auc}}\}$, the update is accepted if $m_k \geq m_{\text{threshold}}$. This approach suits large evaluation datasets where point estimates have low variance.
For production deployments with smaller post-market samples, where point estimates carry greater variance, \gls{ci}-based non-inferiority testing replaces the deterministic comparisons. The deterministic approach used throughout this proof-of-concept is appropriate given the large evaluation dataset; \gls{ci}-based testing is recommended when post-market sample size falls below approximately 500, where bootstrap resampling provides materially stronger guarantees against type I acceptance errors.

\paragraph{\gls{ci}-based equivalence testing (recommended for production)} For smaller post-market samples, \gls{ci} testing provides stronger guarantees. 
Let $\Delta m_k = m_k - m_0$ denote the difference between current and reference performance. Using bootstrap resampling, we compute a $(1-\alpha)$ \gls{ci}. The update is \textbf{non-inferior} if $\mathrm{CI}_{\Delta m_k}^{\mathrm{lower}} \ge -\CImarging$, or \textbf{equivalent} if the interval lies within $[-\CImarging, +\CImarging]$. Margins $\CImarging$ are predefined through clinical risk assessment.

Gold-standard acceptance is evaluated jointly with absolute safety thresholds. Failure of either triggers escalation per the \gls{cdm} hierarchy.

\subsection{\gls{aegis} Overview} 
\begin{figure}[h!]
 \centering
 \includegraphics[width=0.8\textwidth]{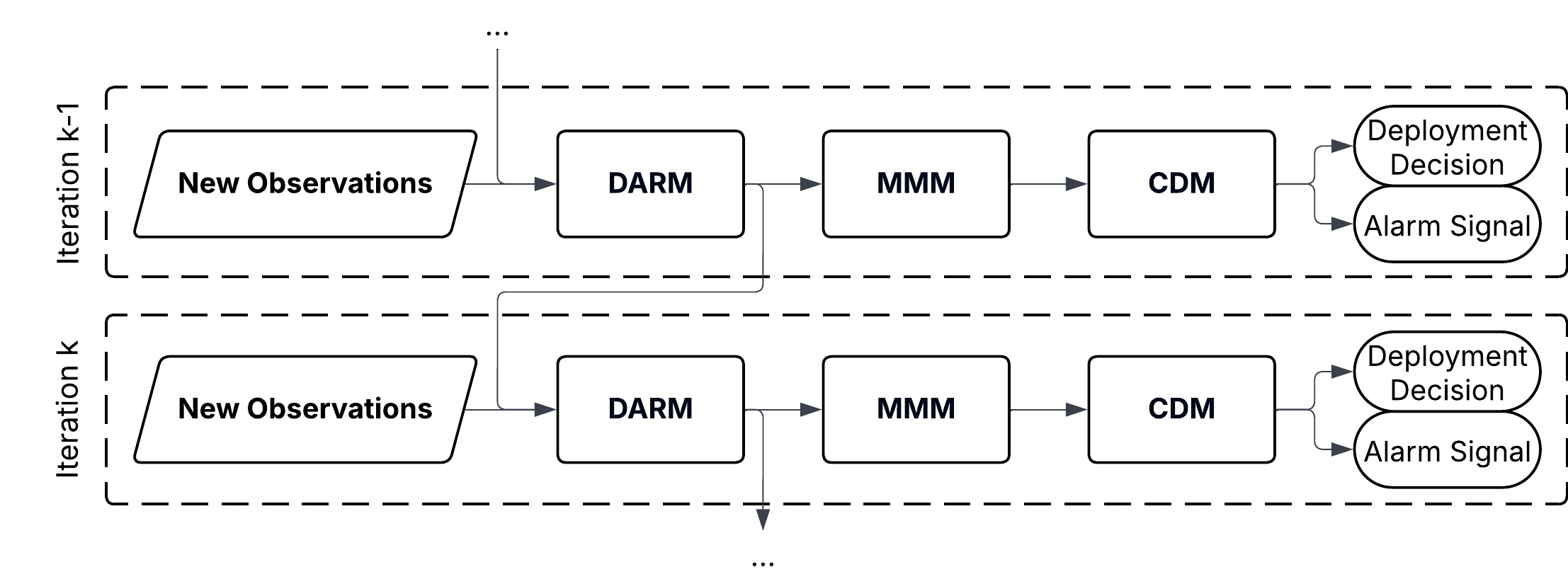}
 \caption{\gls{aegis} \pccpConcept integrating \gls{darm}, \gls{mmm}, and \gls{cdm} at each iteration $k$.}
 \label{fig:model_framework_diagram}
\end{figure}

\noindent
\gls{aegis} \pccpConcept integrates the \gls{darm}, \gls{mmm}, and \gls{cdm} for transparent, automated, auditable model lifecycle management. 
Figure \ref{fig:model_framework_diagram} illustrates the complete \gls{aegis} \pccpConcept.

\gls{aegis} bridges both regulatory paradigms from Table \ref{tab:regulatory_philosophy}. 
The \fixedPerfRef comparison mechanism is designed to support \gls{fda} equivalence documentation. The \gls{mmm}'s continuous monitoring is structured to align with \gls{mdr} surveillance mandates. The four-category deployment taxonomy maps to both: \approve aligns with \gls{pccp} scope and \gls{eu} Article 43(4) exemptions; \reject triggers escalation under each jurisdiction. The parallel \alarm signal addresses \gls{pms} obligations independently of the deployment decision.
\subsection{\gls{aegis} Instantiation}
\label{sec:aegisInstantiation}
\noindent
\gls{aegis} is domain-agnostic: the core modules define \textit{what} to monitor and govern, while \textbf{instantiation parameters} define \textit{how} they operate in specific contexts. Configuration includes: data modality, model architecture, primary metrics, \gls{mlcps} weighting, drift detection method, clinical thresholds, and safety action mappings.

Table \ref{tab:instantiation_config} summarizes configuration for two proof-of-concept instantiations, and the pattern is straightforward: identical governance modules with different configuration parameters. 

Configuration also includes specifying which decision categories are active for a given deployment context. Not all four deployment decision categories need to be enabled in every instantiation; which categories are appropriate depends on the clinical risk profile and regulatory pathway. 
For a high-risk autonomous system, \condApprvl may be clinically inappropriate, i.e., there is no safe middle ground of `deploy with enhanced monitoring' when the cost of an error is severe, so the instantiation would use only \reject, \clinRev, and \approve. 
For a low-risk screening tool with a large operator monitoring infrastructure, \clinRev might be collapsed into \condApprvl. 
Conversely, a system subject to strict regulatory oversight might require that any deviation from \approve triggers \clinRev rather than \condApprvl. 
The active categories and their trigger thresholds must be documented in the \gls{pccp} and derived through \thrSource. 
This configurability is a feature, not a limitation: \gls{aegis} provides the category definitions and priority logic; the instantiation selects the subset appropriate to the clinical context.

\begin{table*}[t]
    \small
    \centering
    \caption{\gls{aegis} Instantiation Parameters}
    \label{tab:instantiation_config}    
    \begin{tabularx}{\textwidth}{@{} m{2.4cm} m{3cm} X X @{}} 
        \toprule
        \multicolumn{1}{c}{\textbf{Parameter}} & \multicolumn{1}{c}{\textbf{\gls{aegis} Definition}} & \multicolumn{1}{c}{\textbf{Sepsis Instantiation}} & \multicolumn{1}{c}{\textbf{Segmentation Instantiation}} \\
        \midrule
        Data modality & Any supervised learning data & Tabular \gls{ehr} time-series (202 features) & 3D multi-parametric \gls{mri} \\
        \midrule
        Model architecture & Agnostic (wrapper) & \gls{rf} & nnU-Net ResENC-M \\
        \midrule
        Primary metrics & Domain-configurable & Sens., Spec., \gls{roc-auc} & \gls{dsc} (\gls{tc}) \\
        \midrule
        \gls{mlcps} weighting & User-defined via risk assessment & Sens.: 1.5, AUC: 1.3, Bal. Acc.: 1.1, Spec.: 1.0 (raw; normalized internally) & \gls{dsc}: 1 \\
        \midrule
        Drift detection & Pluggable method & \BonferroniDriftScoreDef & Performance comparison \\
        \midrule
        Clinical thresholds & Risk-assessment derived (\isoRisk) & Sens. $<$ 0.65$^*$ & \gls{dsc} $<$ 0.676 (\gls{tc})$^{**}$ \\ 
        \midrule
        Safety action & \alarm/\reject triggers & Same \gls{cdm} logic & Same \gls{cdm} logic \\
        \bottomrule
    \end{tabularx}
    \vspace{0.3em}
    \begin{flushleft}
        \footnotesize{Abbreviations: Sens. = sensitivity; Spec. = specificity; Bal. Acc. = balanced accuracy.}\\
        \footnotesize{$^*$\PfailSaftyFloor (\reject triggers). See Table \ref{tab:decision_hierarchy} for complete hierarchy.}\\
        \footnotesize{$^{**}$the \PfailSaftyFloor ($\Pfail = 0.676$) is derived from the \benchmarkModelPerf, defined as the \fixedPerfRef ($\Pref = 0.726$) minus a tolerance $\tauval = 0.05$ (see Table \ref{tab:bt_param_mapping}).}\\
        \footnotesize{Note: \textit{\thrDisclaimer}}
    \end{flushleft}
\end{table*}

\begin{figure}[h!]
 \centering
 \includegraphics[width=0.7\textwidth]{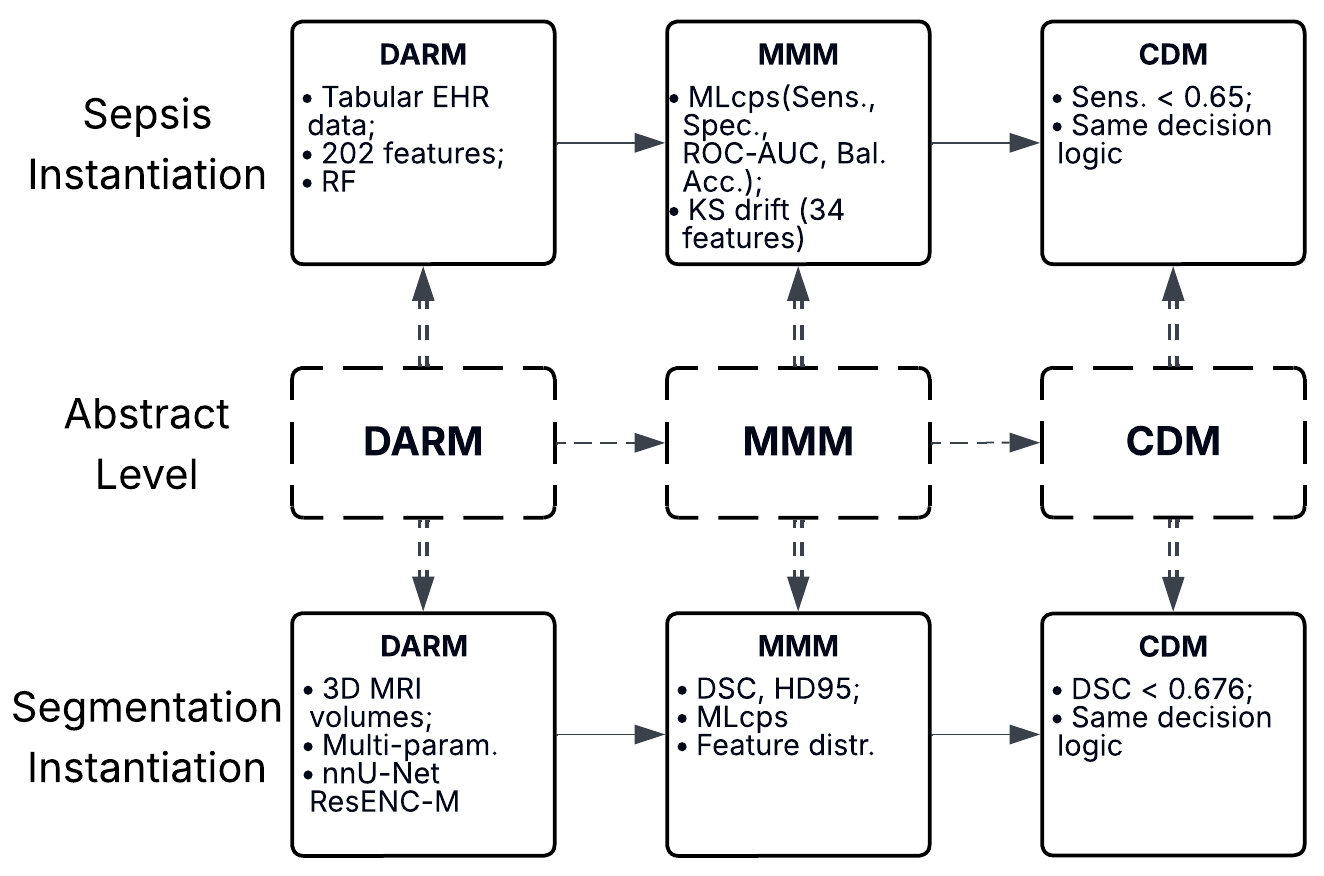}
 \caption{\gls{aegis} instantiation pattern. The abstract \pccpConcept (middle) defines domain-agnostic modules. Each instantiation (top and bottom) configures with domain-specific parameters while maintaining identical logic.}
 \label{fig:aegis_instantiation}
\end{figure}

Figure \ref{fig:aegis_instantiation} illustrates this relationship.

\subsubsection{Proof-of-Concept Validation}
\noindent
We instantiated \gls{aegis} for two distinct applications to demonstrate generality. \textbf{This validation is a proof-of-concept demonstration, not clinical evaluation.} The models serve as vehicles to exercise governance architecture; they have not undergone clinical validation.

The two domains were chosen for complementary characteristics: sepsis prediction (structured time-series with established thresholds) and image segmentation (high-dimensional spatial data with different metrics). \textbf{The goal is demonstrating \gls{aegis} generalizability, not achieving \gls{sota} performance.} Models use conventional architectures without extensive optimization.

Validation encompasses:
\begin{itemize}
 \item Iterative development simulating continuous learning scenarios
 \item \gls{pccp} enforcement through automatic safety checks
 \item \gls{pms} and drift detection
\end{itemize}

\subsubsection{Sepsis Prediction Instantiation}
\noindent
The first instantiation addresses early sepsis prediction using PhysioNet/Computing in Cardiology Challenge 2019 data \cite{reyna_early_2020}. We selected this dataset for accessibility: unlike MIMIC-III/IV (requiring credentialing and data use agreements), PhysioNet data are openly available, supporting reproducibility. 

\textit{Data Source and Preprocessing:} 
Raw data comprised longitudinal records of vital signs (heart rate, pulse oximetry, temperature, blood pressures, respiration rate, end-tidal CO$_2$) and laboratory values (base excess, bicarbonate, pH, BUN, lactate, leukocytes, fibrinogen, platelets). Feature engineering aggregated time-series into fixed-length vectors using statistical features (mean, standard deviation, minimum, maximum, last value, slope) over each ICU stay. Demographics and length of stay were included, yielding 202 features. 
Missing values were median-imputed. 
Features were not standardized, as the \gls{rf} classifier is invariant to monotonic feature transforms.

\textbf{Prediction time horizon.} Features are computed over each patient's complete ICU stay; labels indicate whether sepsis occurred during that stay. This retrospective formulation serves the governance demonstration. Clinical deployment requires specifying:
\begin{itemize}
    \item prediction time relative to admission, 
    \item available measurement window, and 
    \item label definition relative to prediction time.
\end{itemize}
\textit{Model Architecture:}
We employed a single \gls{rf} classifier \cite{breiman_random_2001} (300 estimators, max depth 12, min samples split 10, min samples leaf 5, balanced class weighting, \gls{oob} scoring enabled). These hyperparameters were selected for inter-iteration stability rather than maximum predictive power: the governance demonstration requires that metric changes reflect data perturbations, not model instability. Threshold optimization used \gls{oob} predictions targeting 75\% sensitivity, reflecting clinical priority of detecting true sepsis cases. 
\gls{oob} predictions provide honest, cross-validated probability estimates without requiring a separate validation fold. Features were median-imputed but not standardized, as \glspl{rf} are invariant to monotonic feature transforms. The model was chosen as a conventional baseline; \gls{aegis} is architecture-agnostic.

\textit{\gls{darm} Implementation:} 
The \gls{darm} enforced patient-safe data separation, ensuring all samples from a unique patient ID remained in a single split.
The dataset comprised 20,336 patients from Hospital A of the PhysioNet 2019 Challenge.
The Golden dataset (20\% of patients: 4,067 samples) was reserved for unbiased evaluation. 
The remaining 80\% formed the dynamic pool: half (8,134 samples) for the initial model, half for iterative batches. 
Data accumulation ($\trainDataNext = \trainData \cup \driftData$) is unconditional; it occurs at every iteration regardless of the \gls{cdm} decision. The governance decision controls what happens to the model, not what happens to the data.

\textit{\gls{mmm} Implementation:} 
The \gls{mmm} tracked three metric categories: model performance (accuracy, \gls{roc-auc}, \gls{pr-auc}, F1, sensitivity, specificity, \gls{ppv}, \gls{npv}, \gls{fnr}, \gls{fpr}), fairness (bias score comparing demographic groups), and distribution (drift score).

For unified assessment, we adopted the \gls{mlcps} metric \cite{akshay_mlcps_2023}, which projects multiple metrics onto polar coordinates and calculates enclosed area. 
We configured raw weights: Sensitivity 1.5, \gls{roc-auc} 1.3, Balanced accuracy 1.1, Specificity 1.0. The official \gls{mlcps} package normalizes these internally to proportional angular sectors. These weights are illustrative; production implementations should derive weights through stakeholder consensus with \isoRisk documentation.

\textbf{Drift detection specification.} 
Feature-level drift is assessed via two-sample \gls{ks} tests across monitored features, with Bonferroni correction applied to control the family-wise error rate.

The aggregate drift score $\driftScore$ represents the proportion of features exhibiting statistically significant distributional shift. Minor drift is flagged when $\driftScore$ falls in [0.30, 0.70]; major drift when it exceeds 0.90. 

The drift score quantifies covariate drift, i.e., changes in the marginal distribution of input features. 
Concept drift (shifts in the conditional distribution, i.e., changes in the feature-outcome relationship) is partially addressed through a \textit{proxy} check at Priority 2 (see Table \ref{tab:decision_hierarchy}), which identifies cases where feature distributions appear stable but the released model degrades on new data, a pattern consistent with concept drift, though not conclusive evidence of it without outcome label monitoring. 
Direct concept drift detection, including explicit modelling of the feature-outcome relationship over time, is identified as a future extension (Section \ref{sec:future}).

Clinical threshold estimates are informed by performance bounds observed in published systematic reviews \cite{chua_early_2024, wang_methodological_2025}; for instance, setting a minimum sensitivity floor of $0.65$ aligns with the aggregate performance reported in \cite{chua_early_2024, wang_methodological_2025}.
Additional thresholds include specificity of $0.55$, \gls{roc-auc} of $0.70$, \gls{mlcps} of $0.65$, \gls{npv} of $0.90$, and a maximum \gls{fnr} of $0.35$. 
These values serve as demonstration anchors calibrated to published sepsis systematic reviews \cite{chua_early_2024, wang_methodological_2025}; their purpose is to exercise the governance machinery and verify that the \gls{cdm} routes correctly under each threshold configuration, not to propose universally valid operating values. All threshold values used in this proof-of-concept are illustrative; production implementations must derive each value through formal \isoRisk hazard analysis, with traceability to specific hazard identifiers documented in Supplementary S2.3.

Explainability was assessed using \gls{shap} analysis \cite{lundberg_unified_2017}, quantifying feature importance concentration in top features.

\textit{\gls{cdm} Implementation:}
The \gls{cdm} implemented the four deployment decision categories and parallel \gls{pms} \alarm signal, aligned with US and \gls{eu} regulatory guidance:

\begin{table*}[t]
    \small
    \centering
    \caption{Decision Taxonomy with Regulatory Mapping$^{*}$}
    \label{tab:regulatory_mapping}
    \begin{tabularx}{\textwidth}{@{} p{2.2cm} X X X X @{}}
        \toprule
        \multicolumn{1}{c}{\textbf{Decision}} & \multicolumn{1}{c}{\textbf{Definition}} & \multicolumn{1}{c}{\textbf{\gls{fda} \gls{pccp} Status}} & \multicolumn{1}{c}{\textbf{\gls{eu} \gls{mdr}/\gls{ai} Act}} & \multicolumn{1}{c}{\textbf{Trigger}} \\
        \midrule
        \approve & All metrics satisfied & Within \gls{pccp} scope & Art.\ 43(4) exemption & All pass \\
        \midrule
        \condApprvl & Enhanced monitoring & Limited \gls{pccp} scope & Enhanced \gls{pms} & Minor drift or low trust scores \\
        \midrule
        \clinRev & Human validation required & May exceed \gls{pccp} & Pre-modification review & Near threshold \\
        \midrule
        \alarm & Released model at risk in field; issued as a \gls{pms} signal independently of the new model deployment decision & Activates \gls{pms} reporting; may require vigilance report depending on severity & \gls{eu} \gls{mdr} Art.\ 83 \gls{pms} obligation; potential Art.\ 87 serious incident notification & \PPMSSaftyFloor breach; released model regression; major distributional shift \\
        \midrule
        \reject & Deployment blocked & New 510(k) required & New \gls{ca} & Safety violation \\
        \bottomrule
    \end{tabularx}
    \begin{flushleft}
        \footnotesize{$^{*}$ \textit{Regulatory status mappings represent operational analogies intended to guide \gls{pccp} documentation structure; definitive regulatory classification depends on device-specific intended use, risk classification, and jurisdictional interpretation, and should be confirmed with regulatory counsel prior to any submission}.}\\
        \footnotesize{Note: \alarm is a parallel \gls{pms} signal and is not part of the deployment decision hierarchy. It may be co-issued with any deployment decision. When \alarm is absent, no label is emitted. Regulatory reporting obligations triggered by \alarm depend on device classification, severity assessment, and jurisdictional requirements; manufacturers should document thresholds and escalation pathways in their \gls{pms} plan.}
    \end{flushleft}    
\end{table*}

The decision hierarchy evaluates conditions in priority order (Table \ref{tab:decision_hierarchy}). This ensures safety violations are detected first.

\textbf{Two-tier threshold architecture.} \PfailSaftyFloor defines floors below which deployment is prohibited (\reject). Performance thresholds define acceptable bounds; falling below triggers \clinRev. This buffer ensures human oversight before rejection.

\textbf{\gls{cdm} specification (deterministic implementation):}

To ensure complete auditability and reproducibility, the \gls{cdm} is specified as a deterministic function:

\noindent\textbf{Inputs:} Metrics on $\goldData$ and $\driftData$ (sensitivity, specificity, \gls{roc-auc}, \gls{npv}, \gls{mlcps}), drift indicators (minor/major flags, drift score), trustworthy \gls{ai} indicators (bias score, explainability score), and gold-standard acceptance results (threshold comparisons or \gls{ci}-based test outcomes).

\noindent\textbf{Outputs:} \{decision\_label, trigger\_reasons[], required\_actions[], logged\_artifacts[]\}.

\noindent\textbf{Logging requirements:} For audit trail completeness, the \gls{cdm} must record: metric values at decision time, threshold/margin values used, dataset identifiers and hashes, drift statistics with p-values, and explicit trigger reasons. This logging enables regulatory review of every governance decision.

\noindent\textbf{Priority logic:}
\begin{enumerate}
    \item If the \PgCurDef is less than the \PfailSaftyFloor ($\Pcur <  \Pfail = 0.65$) $\rightarrow$ \textbf{\reject} \textnormal{[P1]} (reason: \PfailSaftyFloor violation)
    \item Else if any metric in buffer zone (sensitivity $< 0.66$ OR specificity $< 0.60$) OR the regression of the \PgCurDef from the \fixedPerfRefLong is sufficiently large ($\tauval < \Pref - \Pcur$) $\rightarrow$ \textbf{\clinRev} \textnormal{[P2]} (reason: borderline safety/performance)
    \item Else if minor drift ($0.30 \leq \driftScore < 0.70$) OR \gls{tai} concern $\rightarrow$ \textbf{\condApprvl} \textnormal{[P3]} (reason: enhanced monitoring required)
    \item Else $\rightarrow$ \textbf{\approve} \textnormal{[P4]} (reason: all criteria satisfied)
\end{enumerate}
\noindent\textit{\gls{pms} \alarm signal (evaluated in parallel using released model on $\driftData$; fires if any condition is true):}
\begin{enumerate}
    \item[A1)] The \PdRelDef is less than the \PPMSSaftyFloor ($\PdRel < \PPMS$) $\rightarrow$ \textbf{\alarm} (reason: \PPMSSaftyFloor breached)
    \item[A2)] The \PdRelDef is less than the \PgRelDefLong by more than a tolerance ($\PdRel \leq \PgRel - \tauval$) $\rightarrow$ \textbf{\alarm} (reason: released model regressed from the reference on new data)
    \item[A3)] If the \BonferroniDriftScoreDef is sufficiently large ($0.90 < \driftScore$) $\rightarrow$ \textbf{\alarm} (reason: major distributional shift threatens released model validity)
\end{enumerate}
The composite output is reported as (deployment decision) or (deployment decision + \alarm) when any of A1--A3 fires. Absence of \alarm requires no label.

Complete algorithmic pseudo-code and implementation guidance appear in Supplementary Materials S3.3.

\textbf{Implementation note on acceptance criteria.} This proof-of-concept employs deterministic threshold comparisons (e.g., sensitivity $\geq$ 0.65) rather than \gls{ci}-based equivalence testing. This choice reflects the pedagogical goal of demonstrating core functionality with maximum interpretability, appropriate given the large evaluation dataset ($N = 4{,}067$ patients). For production deployments with smaller post-market samples, \gls{ci}-based non-inferiority testing is recommended.

\textbf{\makefirstuc{\fixedPerfRef} comparison coupling.} The \fixedPerfRef comparison mechanism serves two distinct functions in the composite \gls{cdm} output. First, it acts as a trigger for the deployment decision at P2 (\clinRev): if the \PgCurDef ($\Pcur$) has regressed from the \fixedPerfRefLong ($\Pref$) by more than a tolerance ($\tauval$), a \clinRev is triggered even when absolute metrics remain above the \PfailSaftyFloor. Second, it provides the reference (i.e., $\PgRel$, which is the \PgRelDefLong) against which the \gls{pms} \alarm trigger A2 is evaluated, i.e., if the \PdRelDef ($\PdRel$) has degraded by more than a tolerance ($\tauval$) from its reference ($\PgRel$), \alarm fires. 
Upon initial \approve, \gls{aegis} stores validated metrics both as $\Pref$ and as $\PgRel$ for reference checks. Subsequent \approve fires update $\PgRel$ to the newly released model's performance on the Golden set. 

\textbf{$\Pref$ \textit{vs} $\PgRel$ distinction.} The \fixedPerfRefLong ($\Pref$) is fixed at the first \approve and never updated, while the \PgRelDefLong ($\PgRel$) is a rolling reference that advances with each subsequent \approve. 
The \glsposs{aegis} dual-criterion approach ensures that the \PgCurDef ($\Pcur$) does not silently degrade below the \PgRelDef ($\PgRel$) or the \fixedPerfRef ($\Pref$).

\textbf{Minor drift configuration.} The minor drift range (0.30--0.70) in Table \ref{tab:decision_hierarchy} represents a configurable threshold for triggering enhanced monitoring via \condApprvl. In this validation, all four deployment decision categories are enabled and exercised. The minor drift range was widened from [0.35, 0.70] to [0.30, 0.70] based on threshold calibration analysis of stationary baseline variability (iterations 0--5), consistent with \isoRisk principles of deriving thresholds from observed operational characteristics.
The complete mapping of abstract \gls{aegis} parameters to specific values and clinical justifications for the sepsis prediction instantiation is summarized in Table \ref{tab:sepsis_param_mapping}.

\begin{table*}[!ht]
    \small
    \centering
    \caption{Mapping of Abstract \gls{aegis} Parameters to Sepsis Prediction Parameters.
    This table provides the sepsis-specific instantiation of the \pccpConcept-level
    \gls{cdm} priority hierarchy defined in Table~\ref{tab:decision_hierarchy}.}
    \label{tab:sepsis_param_mapping}
    \begin{tabularx}{\textwidth}{@{} m{3cm} m{4cm} X m{3cm} @{}}
        \toprule
        \multicolumn{1}{c}{\textbf{\gls{aegis} Parameter}} & \multicolumn{1}{c}{\textbf{Concrete Value}} & \multicolumn{1}{c}{\textbf{Clinical Justification}} & \multicolumn{1}{c}{\textbf{\gls{cdm} Decision}} \\
        \midrule
        $\Pref$ & Sens.: 0.723, Spec.: 0.933, AUC: 0.922 & \fixedPerfRefLong & \fixedPerfRef \\
        \midrule
        $\Pfail$ & Sens.: $0.65$ & derived from published systematic reviews of sepsis prediction performance \cite{chua_early_2024, wang_methodological_2025} & \reject \\
        \midrule
        $\RaccG$ &  $0.66 \leq$ Sens.\, $0.60 \leq$ Spec.& Performance thresholds (buffer zone) & \clinRev \\
        \midrule
        Major drift threshold & $0.90 < \driftScore$ & $>90\%$ of 34 key features shifted (\gls{ks} test) & \alarm (A3) \\
        \midrule
        Minor drift range & $0.30 \leq \driftScore \leq 0.70$ & Moderate distributional change detected & \condApprvl \\
        \midrule
        $\PPMS^*$ & Sens.: $0.65$ &  $=\Pfail$ & \alarm (A1) \\
        \midrule
        $\tauval$ (\gls{pms} regression margin) & $0.020$--$0.030$ sens. points & Detects gradual decline before \PPMSSaftyFloor is breached & \alarm (A2) \\
        \midrule
        $\Pcur$, $\PDcur$  & \gls{mlcps}(Sens.: 1.5, AUC: 1.3, Bal. Acc.: 1.1, Spec.: 1.0; raw weights, normalized internally) & Sensitivity-weighted per clinical priority & Composite tracking \\
        \midrule
        \gls{npv} floor & $0.90 \leq$ & Safe negative triage in screening context & Safety constraint \\
        \midrule
        \gls{fnr} ceiling & $\leq 0.35$ & Maximum tolerable missed sepsis cases & Safety constraint \\
        \midrule
        --- & Model architecture & Single \gls{rf} (300 trees, depth 12, \gls{oob} threshold) & Architecture specification \\
        \midrule
        \fixedPerfRefTol & 0.015 (sensitivity) & Detects subtle regression ($>$ 1.5\% decline from \fixedPerfRef) & \clinRev \\
        \midrule
        $\lvert \goldData \rvert$ & $N = 4{,}067$ (20\% of 20,336) & Sufficient for deterministic thresholds & Evaluation dataset \\
        \bottomrule
    \end{tabularx}
    \vspace{0.3em}
    \begin{flushleft}
        \footnotesize{Abbreviations: Sens. = sensitivity; Spec. = specificity; Bal. Acc. = balanced accuracy.}\\
        \footnotesize{Notations: $\Pref$ = \fixedPerfRef; $\tauval$ = \tolDef; $\Pfail/\PPMS$ = \PfailPMS; $\RaccG$ = acceptable performance ranges for Golden dataset; $\Pcur/\PDcur$ = \PgoldDrift; $\driftScore$ = \BonferroniDriftScoreDef; $\lvert \goldData \rvert$ = Golden set size.}\\  
        \footnotesize{$^*$In this proof-of-concept, $\PPMS$ is set equal to $\Pfail$ for simplicity. \PpmsPfailDistinction}\\
        \footnotesize{Note: \textit{\thrDisclaimer}}
    \end{flushleft}
\end{table*}

\textit{Iterative Simulation:}
The simulation ran 11 iteration cycles, systematically exercising the governance decision space. The model at iteration 0 trained on 8,134 samples (714 sepsis cases, 8.8\%). Iterations 1--5 used real data batches ($\sim$1,626 samples each) from the iterative pool. Iterations 6--10 used mixed and synthetic batches designed to exercise each of the four deployment decision categories and both \alarm composite states:

\begin{itemize}
    \item 
    \textbf{Cross-site minor drift} (iteration 6): 50\% Hospital A + 50\% Hospital B patients (cross-hospital distributional difference); targets \condApprvl via minor drift detection
    \item 
    \textbf{Regression from \fixedPerfRef $\Pref$} (iteration 7): Clean Hospital A batch (n=3,000). Cross-site contamination from iteration 6 causes subtle sensitivity regression detected by \fixedPerfRef comparison ($>$1.5\% decline from \fixedPerfRef); targets \clinRev
    \item 
    \textbf{Extreme distributional shift} (iteration 8): 5$\times$ scaling + 5$\times\sigma$ offset on all 34 drift features (n=100, small batch to avoid model corruption); targets \approve + \alarm (new model unaffected by small batch; \gls{pms} \alarm fires via A3 major drift)
    \item 
    \textbf{Recovery} (iteration 9): Clean Hospital A resample (n=3,000); demonstrates system recovery after drift/regression; targets \approve
    \item 
    \textbf{Catastrophic corruption} (iteration 10): 80\% positive label flip, 30\% negative label flip, $\sigma$=4$\times$std Gaussian noise on all features (n=25,000); targets \reject + \alarm (new model sensitivity collapses below 0.65; released model simultaneously fails on same corrupt batch)
\end{itemize}

The specific perturbation magnitudes are illustrative rather than calibrated to real-world drift characteristics. While synthetic scenarios enable controlled evaluation, they may not fully capture real-world complexity from demographic changes, equipment variations, or clinical practice evolution \cite{sahiner_data_2023}. Future validation should incorporate documented real-world drift events.

\subsubsection{Brain Tumor Segmentation Instantiation}
\label{sec:BrainTumorSegInst}
\noindent
The second instantiation demonstrates the \glsposs{aegis} utility in the domain of medical imaging, specifically targeting the automated segmentation of brain tumors in \gls{mri} scans. 
Radiology remains a cornerstone for clinical \gls{ai} applications, representing one of the most established fields for \gls{ml} integration in healthcare. 
To ensure a robust evaluation, we utilized data and task definitions from the \gls{miccai} \gls{brats} challenge \cite{menze_multimodal_2015, bakas_identifying_2019}. 
This benchmark was selected due to its long-standing maturity and the availability of extensively validated methods, providing a standardized environment to test \glsposs{aegis} ability to govern high-dimensional spatial data.

\textit{Dataset and the studied task:} 
This study utilizes data from the \gls{brats} 2023 and 2024 challenges, focusing on the pre-operative and post-treatment \gls{gli} cohorts. 
The dataset comprises \gls{mpmri} scans, including \gls{t1n}, \gls{t1c}, \gls{t2w}, and \gls{t2f} sequences. 
These scans are paired with expert-validated segmentation masks that delineate distinct tumor subregions.
In pre-operative cases, these masks identify the \gls{et}, \gls{netc}, and peritumoral \gls{ed}. For post-treatment scans, the segmentation task is extended to include the \gls{rc}. 

\begin{figure}[h!]
 \centering
 \includegraphics[width=0.8\linewidth]{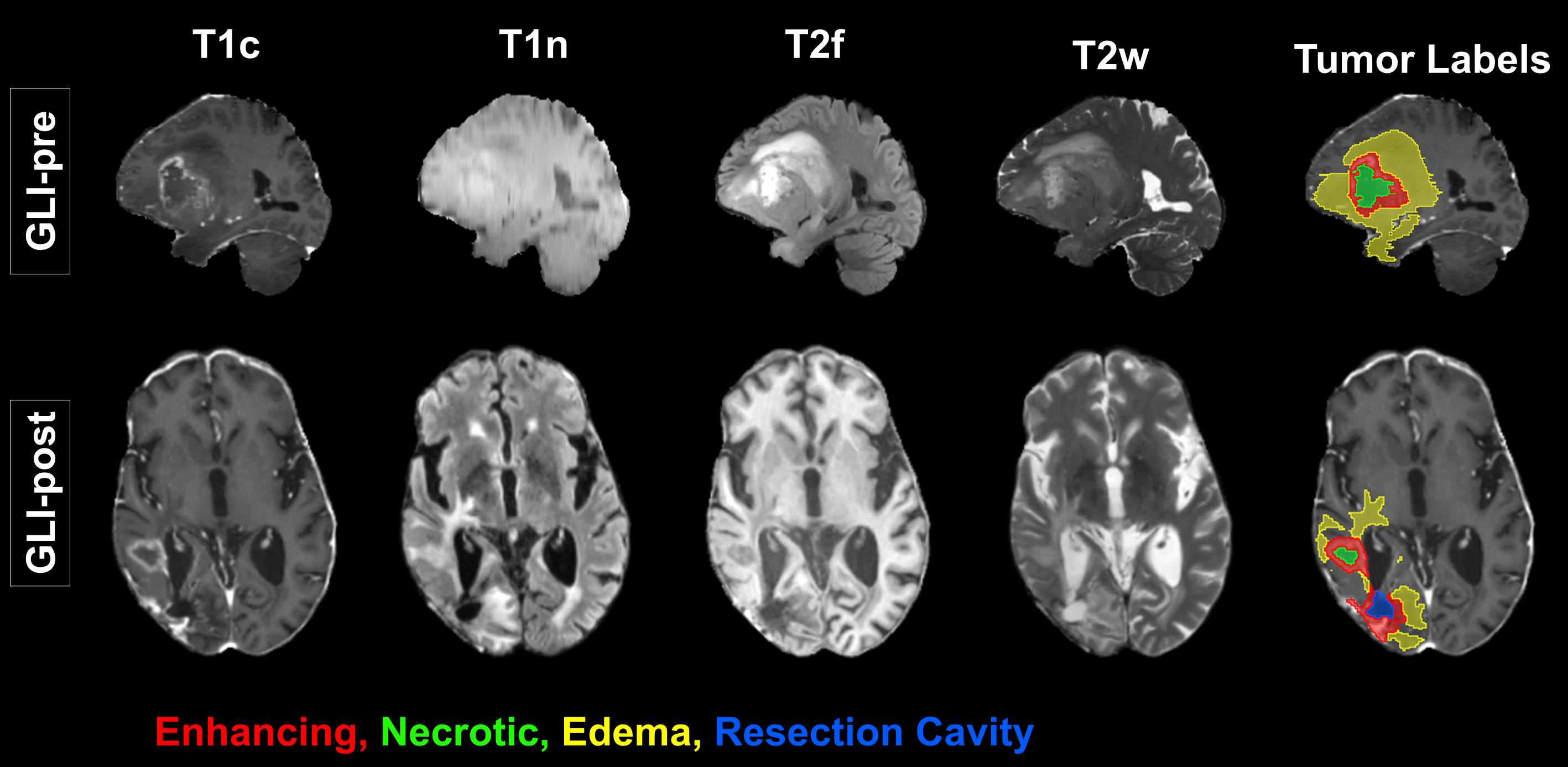}
 \caption{\gls{gli} tumors in pre-operative and post-treatment \gls{mpmri}.}
 \label{fig:context-brain}
\end{figure}

Figure \ref{fig:context-brain} illustrates the complexity of this task by visualizing the heterogeneous appearance of these tumor subregions across both pre-operative and post-treatment \gls{mpmri} modalities.

\textit{Tumor segmentation workflow}
To operationalize \gls{aegis} for volumetric data, we utilized the nnU-Net ResENC-M architecture \cite{isensee_nnu-net_2021}, version 2 (nnU-Net v2)---a highly robust solution that has consistently secured winning positions in several \gls{brats} challenges. 
The model was implemented using its default configurations to ensure a standardized and reproducible environment for evaluating the \gls{aegis} governance modules. 
The network accepted the four \gls{mpmri} sequences as multi-channel inputs and underwent training for 1,500 epochs for direct label-wise segmentations.
While various quantitative metrics are common to evaluate segmentation performance, we chose to use the \gls{dsc} due to its prevalence and conventional use in the medical image segmentation community.

Although the model was trained to segment all constituent subregions, \gls{aegis} \gls{mmm} focused its governance logic specifically on the TC, which is the composite region of the \gls{et} and \gls{netc}. 
This subregion was prioritized for two primary reasons. First, the \gls{tc} holds important clinical significance, as it is the primary target for radiotherapy and surgical planning; therefore, its accuracy is the most critical metric for safety-centric governance. 
Second, focusing on a consolidated clinical target streamlines the statistical analysis for drift detection and the subsequent decision-making process, including automated acceptance or rejection, ensuring \gls{aegis} remains computationally efficient and clinically interpretable.

\textit{Iterative simulations}
The evaluation began by establishing a \fixedPerfRef using a held-out \textit{Golden dataset} ($\goldData$) consisting of 200 subjects from the \gls{brats} post-treatment \gls{gli} cohort.
The corresponding model was trained on the remaining 1,150 subjects, and its performance on the $\goldData$ was recorded as the \fixedPerfRef.

To simulate a continuous learning lifecycle, we implemented an iterative training scheme.
The process was initialized by training a model on a small subset of 200 subjects from the post-treatment training pool. 
In each subsequent iteration, a new Drifting Dataset was introduced to the \gls{mmm} to assess real-time stability. 
While metrics from the drifting data did not dictate the final release status, significant deviations from the Golden Dataset performance served as early warnings for potential catastrophic forgetting or distribution shifts. 
Following each evaluation, the new data was assimilated into the training pool for the next iteration, reflecting a continuous data-acquisition and retraining pipeline.

This iterative process continued until the entire post-treatment cohort was integrated. 
To evaluate the \glsposs{aegis} resilience against significant distribution shifts, we then introduced 1,251 pre-operative datasets. 
We hypothesized that the model, originally trained to segment the \gls{rc}, would maintain its robustness despite the absence of this feature in pre-operative scans. 
The full experimental sequence comprised 12 iterations, as summarized in Table \ref{tab:dice_results}.

To operationalize the \gls{cdm}, we established a deterministic performance threshold based on the \benchmarkModel. 
Following regulatory logic for non-inferiority, the threshold for model approval was set at a 5\% margin below the \fixedPerfRef ($\Pref = 0.726$), yielding $\Pfail = 0.726 - 0.05 = 0.676$. 
A model iteration is only eligible for deployment if its performance on the static Golden Dataset remains above this predefined \PfailSaftyFloor.
Parallel to this, the Drifting Dataset was monitored to detect shifts in the distribution of incoming data batches. 
A decrease in \gls{dsc} on the drifting set relative to the previous iteration serves as a proxy for identifying ``hard'' samples, such as rare tumor morphologies, noisy acquisitions, or expert labeling variability. 
This dual-track monitoring, static (Golden) and dynamic (Drifting), ensures that the model maintains core safety while the system remains sensitive to environmental changes.

The resulting regulatory decision logic for each iteration ($k$) is summarized as follows:
\begin{itemize} 
    \item \textbf{Reject}: if $\Pcur < \Pfail = 0.676$ (Safety violation/Performance failure); OR if $\Pcur < \PgRel$ (Performance degradation below the last \textit{released} version, rather than merely the immediately preceding iteration) 
    \item \textbf{Approve}: if $\PgRel < \Pcur$ AND $\PDprev < \PDcur$ (Performance improvement over \PgRelDef $\PgRel$ with data stability). 
    \item \textbf{Alarm}: if $\Pprev < \Pcur$ AND $\PDcur<\PDprev$ (Golden set recovered, but a potential distribution shift was detected in the new data).   
\end{itemize}
The comprehensive mapping of the abstract \gls{aegis} parameters to the specific configurations used for the brain tumor segmentation instantiation is detailed in Table \ref{tab:bt_param_mapping}.

\textbf{Note on active category configuration:} The \condApprvl and \clinRev categories were intentionally not activated in this brain tumor instantiation. This reflects a deliberate configuration choice for the specific clinical context: brain tumor segmentation for treatment planning is a high-consequence application in which intermediate deployment states carry limited clinical utility, i.e., a model either meets the non-inferiority threshold for treatment planning use or it does not. Activating only \reject, \approve, and \alarm for this instantiation is therefore appropriate, not a validation limitation. This asymmetry between instantiations illustrates the configurability principle described in Section~\ref{sec:aegisInstantiation}: the sepsis instantiation activates all four deployment decision categories to demonstrate the complete decision space, while the brain tumor instantiation activates a clinically appropriate subset. Taken together, the two instantiations establish both that all categories are reachable through configuration and that selective activation is a legitimate design choice. Production implementations of either domain should document the active category set and its clinical justification in the \gls{pccp}.
\section{Results} 
\noindent
The sepsis experiment (Section \ref{sec:SepsisPrediction}) is designed to demonstrate decision-space coverage through calibrated perturbations; it characterises governance mechanism behavior, not real-world drift prevalence or frequency. 
The brain tumor case study (Section \ref{sec:BrTS}) derives governance decisions from genuine distributional shifts across imaging cohorts, providing a complementary naturalistic trajectory. Together, the two instantiations establish that AEGIS produces the correct decision under controlled conditions (sepsis) and under real distributional change (brain tumor), but neither constitutes epidemiological evidence of any drift type’s clinical prevalence.
\subsection{Sepsis Prediction}
\label{sec:SepsisPrediction}
\noindent
\gls{aegis} validation used the PhysioNet Sepsis Challenge 2019 data. Results demonstrate governance behavior, not clinical suitability. The Golden dataset comprised 4,067 samples from 4,067 patients.

\begin{figure}[h!]
 \centering
 \includegraphics[width=1\linewidth]{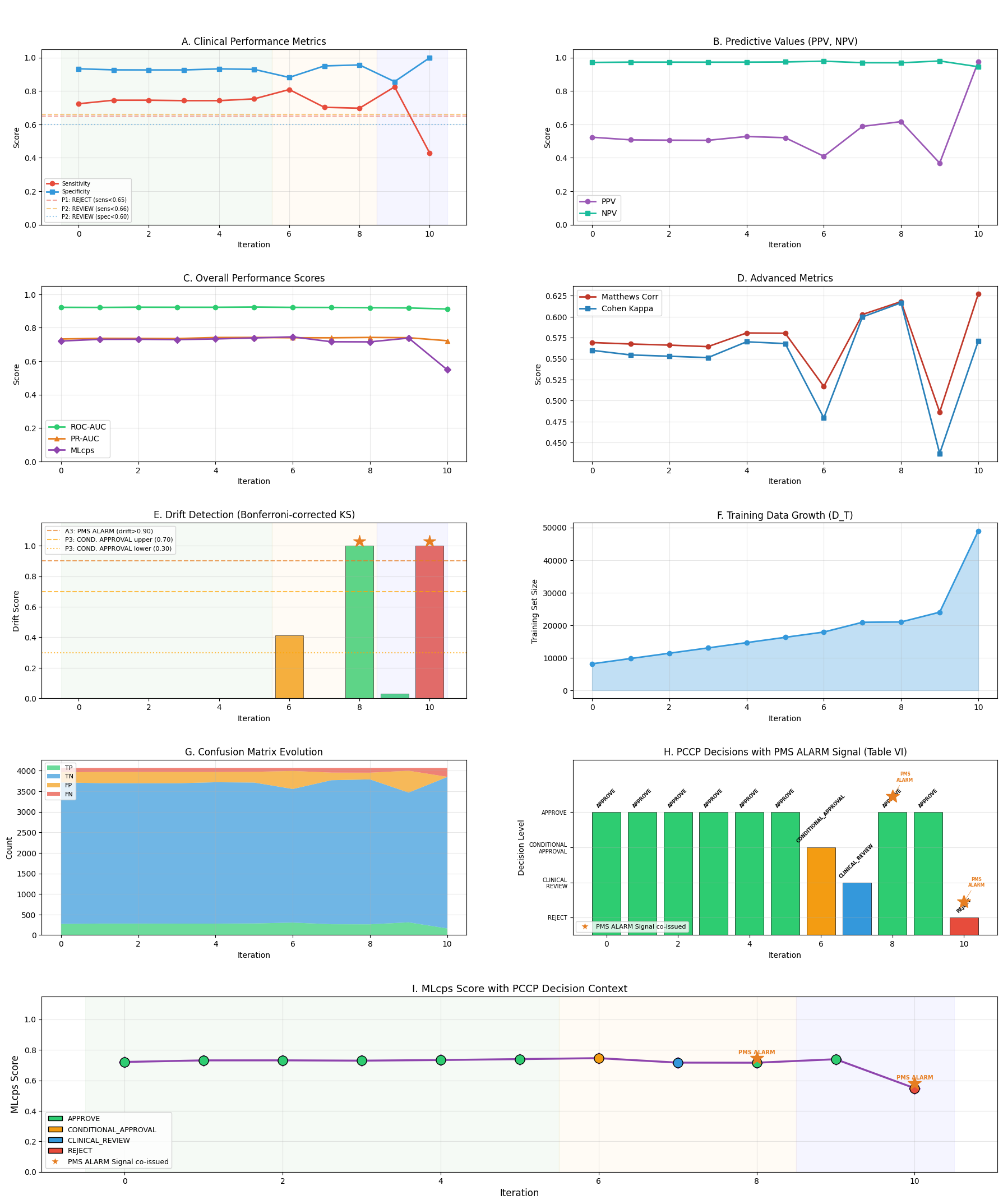}
 \caption{\gls{aegis} comprehensive iterative change control results. A) Clinical performance metrics; B) \gls{ppv} and \gls{npv}; C) \gls{roc-auc}, \gls{pr-auc}, and \gls{mlcps}; D) Matthews correlation and Kappa; E) Drift score; F) Training size; G) Confusion matrix components; H) Deployment decisions with \gls{pms} \alarm markers at iterations 8 and 10; I) \gls{mlcps} score with \gls{cdm} decision context.}
 \label{fig:sepsisResults}
\end{figure}

Figure \ref{fig:sepsisResults} shows results across 11 assessments (iterations 0--10).

The initial model (iteration 0) achieved sensitivity 0.723, specificity 0.933, \gls{roc-auc} 0.922. Iterations 1--5 (real data from Hospital A) maintained stable metrics with small fluctuations from retraining on progressively larger training sets: sensitivity ranged 0.742--0.753, specificity 0.926--0.933. The training dataset grew from 8,134 to 16,268 samples across these iterations. Drift scores remained at 0.000, confirming distributional stability within the same hospital population.

Synthetic scenarios (iterations 6--10) exercised all four deployment decision categories and both \alarm composite states. 
Iteration 6 (cross-site data) produced a drift score of 0.412, triggering \condApprvl. Iteration 7 (clean data, but cross-site contamination in $\trainData$) caused sensitivity to drop to 0.702---above the Priority 2 hard threshold (0.66) but 0.021 below the \fixedPerfRef ($\Pref$), triggering \clinRev via the \fixedPerfRef comparison mechanism. Iteration 8 (extreme feature scaling, n=100) produced drift score 1.000, the small batch left the accumulated training pool intact so the new model passed all P4 checks (sensitivity 0.697), yielding \approve, while the \gls{pms} \alarm fired because the released model failed on the same extreme batch (\approve + \alarm).
Iteration 9 (clean recovery batch) restored all metrics, yielding \approve. Iteration 10 (catastrophic label and feature corruption, n=25,000) collapsed sensitivity to 0.428, triggering \reject; the released model simultaneously failed on the same corrupt batch, firing \alarm (\reject + \alarm, i.e., the critical composite state).

\gls{cdm} deployment decisions: \approve (iterations 0--5 and 9), \condApprvl (iteration 6), \clinRev (iteration 7), \reject (iteration 10). \gls{pms} \alarm signals were co-issued at iteration 8 (\approve + \alarm: new model passed all deployment checks, but the released model was assessed as at risk from the major distributional shift) and iteration 10 (\reject + \alarm: new model unsafe and released model simultaneously failing, i.e., the critical composite state). 
\gls{mlcps} (sensitivity-weighted, computed via the official \gls{mlcps} package) ranged from 0.7207 (iteration 0) to 0.7456 (iteration 6, peak) and collapsed to 0.5485 at iteration 10. \gls{roc-auc} remained stable throughout (0.912--0.924).

Table \ref{tab:iteration_summary} provides comprehensive results. 
The ``Routing Trace'' column in Table \ref{tab:iteration_summary} traces the exact priority-ordered evaluation for the deployment decision at each iteration, serving as a concrete instantiation of the \gls{cdm} logic formalised in Tables \ref{tab:decision_variants} and \ref{tab:decision_hierarchy}.

\begin{table}[ht]
    \small
\centering
\caption{Iteration Summary with Trigger Conditions. The Routing Trace column traces the exact priority-ordered evaluation for the new candidate model at each iteration. The \gls{pms} Signal column records \alarm co-issuances where the released model was assessed as at risk.}
\label{tab:iteration_summary}
\begin{tabularx}{\textwidth}{@{}
>{\centering\arraybackslash}m{.3cm}
>{\centering\arraybackslash}m{.7cm}
>{\centering\arraybackslash}m{.5cm}
>{\centering\arraybackslash}m{.5cm}
>{\centering\arraybackslash}m{.5cm}
>{\centering\arraybackslash}m{.6cm}
>{\centering\arraybackslash}m{0.5cm}
>{\centering\arraybackslash}m{1.8cm}
>{\centering\arraybackslash}m{2.5cm}
>{\centering\arraybackslash}m{.8cm}
>{\centering\arraybackslash}X
@{}}
\toprule
\textbf{Iter.} & \textbf{$\lvert \trainData \rvert$} & \textbf{Sens.} & \textbf{Spec.} & \textbf{AUC} & \textbf{\gls{mlcps}} & \textbf{$\driftScore$} & \textbf{Trigger} & \textbf{Routing Trace$^*$}& \textbf{Deploy. \mbox{Decision}}& \textbf{\gls{pms} Signal} \\
\midrule
0 & 8,134 & 0.723 & 0.933 & 0.922 & 0.721 & 0.000 & All satisfied & \prio{1}$\checkmark\,$ \prio{2}$\checkmark\,$ \prio{3}$\checkmark\,$ $\rightarrow$ \prio{4} & APPR. & ---\\
1 & 9,760 & 0.745 & 0.926 & 0.922 & 0.731 & 0.000 & All satisfied & \prio{1}$\checkmark\,$ \prio{2}$\checkmark\,$ \prio{3}$\checkmark\,$ $\rightarrow$ \prio{4} &APPR. &---\\
2 & 11,386 & 0.745 & 0.926 & 0.923 & 0.731 & 0.000 & All satisfied & \prio{1}$\checkmark\,$ \prio{2}$\checkmark\,$ \prio{3}$\checkmark\,$ $\rightarrow$ \prio{4} &APPR. &---\\
3 & 13,012 & 0.742 & 0.926 & 0.923 & 0.729 & 0.000 & All satisfied & \prio{1}$\checkmark\,$ \prio{2}$\checkmark\,$ \prio{3}$\checkmark\,$ $\rightarrow$ \prio{4} &APPR.  & ---\\
4 & 14,638 & 0.742 & 0.932 & 0.923 & 0.733 & 0.000 & All satisfied & \prio{1}$\checkmark\,$ \prio{2}$\checkmark\,$ \prio{3}$\checkmark\,$ $\rightarrow$ \prio{4} &APPR. & ---\\
5 & 16,268 & 0.753 & 0.929 & 0.924 & 0.739 & 0.000 & All satisfied & \prio{1}$\checkmark\,$ \prio{2}$\checkmark\,$ \prio{3}$\checkmark\,$ $\rightarrow$ \prio{4} &APPR. & ---\\
6 & 17,894 & 0.809 & 0.881 & 0.922 & 0.746 & 0.412 & minor drift $\in$ [0.30, 0.70] & \prio{1}$\checkmark\,$ \prio{2}$\checkmark\,$ $\rightarrow$ \prio{3} & COND.\ APPR. & ---\\
7 & 20,894 & 0.702 & 0.950 & 0.922 & 0.716 & 0.000 & regression from $\Pref$ = -0.021 & \prio{1}$\checkmark\,$ $\rightarrow$ \prio{2} & CLIN.\ REV. & ---\\
8 & 20,994 & 0.697 & 0.956 & 0.920 & 0.716 & 1.000 & $0.90 < \driftScore$ & \prio{1}$\checkmark\,$ \prio{2}$\checkmark\,$ \prio{3}$\checkmark\,$ $\rightarrow$ \prio{4} & APPR. & \alarm \\
9 & 23,994 & 0.824 & 0.856 & 0.919 & 0.739 & 0.029 & All satisfied & \prio{1}$\checkmark\,$ \prio{2}$\checkmark\,$ \prio{3}$\checkmark\,$ $\rightarrow$ \prio{4} & APPR. & ---\\
10 & 48,994 & 0.428 & 0.999 & 0.912 & 0.549 & 1.000 & sens. $<$ 0.65; $0.90 < \driftScore$ & $\rightarrow$ \prio{1} & REJ. & \alarm \\
\bottomrule
\end{tabularx}
\vspace{0.3em}
\begin{flushleft}
    \footnotesize{Abbreviations: Iter. = iteration; Sens. = sensitivity; Spec. = specificity; Deploy. = Deployment; APPR. = \approve; COND.\ APPR. = \condApprvl; CLIN.\ REV. = \clinRev; REJ. = \reject.}\\
    \footnotesize{Notations: $\lvert \trainData \rvert$ = training set size. $\driftScore$ = \BonferroniDriftScoreDef.}\\
\footnotesize{$^*$Uses Table \ref{tab:decision_hierarchy} notation for priorities ($\checkmark$ = check passed; $\rightarrow$ = selected priority).}\\
    \footnotesize{Note: \textit{\thrDisclaimer}}
\end{flushleft}
\end{table}

A key observation: at iteration 8, the new candidate model received an \approve deployment decision (sensitivity 0.697, above all thresholds; the small n=100 batch did not corrupt the accumulated training pool), while a \gls{pms} \alarm was simultaneously co-issued because the released model's performance on the same extreme batch collapsed and the drift score (1.000) exceeded the A3 threshold. This \approve + \alarm composite demonstrates two properties of  \gls{aegis}: (1) drift detection continues to operate as a \textbf{leading indicator}, identifying distributional risk before aggregate performance degradation is reflected in the new model; and (2) deployment and \gls{pms} monitoring are decoupled, i.e., the presence of field risk accelerates the case for deploying the new model rather than blocking it. This proactive approach aligns with \gls{fda} guidance on monitoring ``reasonably foreseeable'' changes.

A second observation: iterations 0--5 show small metric fluctuations (sensitivity 0.723--0.753) despite retraining on progressively larger training sets (8,134 to 16,268 samples). This reflects the stable underlying distribution: batches are drawn from the same hospital population. Drift scores remain at 0.000 during this period, confirming distributional stability. The non-zero variability in sensitivity (unlike the previous implementation's identical metrics) arises because the revised implementation unconditionally accumulates data and retrains at every iteration, correctly implementing the \gls{darm} protocol where $\trainDataNext = \trainData \cup \driftData$.

A third observation: the \fixedPerfRef comparison mechanism (iteration 7) detected a subtle sensitivity regression of 0.021 from the \fixedPerfRef ($\Pref = 0.723$). Although sensitivity (0.702) remained above the Priority 2 hard threshold (0.66), the \fixedPerfRefTol ($\tauval = 0.015$) caught this clinically meaningful decline, demonstrating the two-tier threshold architecture: hard floors for safety, and \fixedPerfRef comparison for quality assurance.

\gls{shap} analysis indicated top features related to vital sign variability and laboratory markers. Bias score (sensitivity difference across demographic-proxy subgroups) remained minimal across iterations.

\subsection{Brain Tumor Segmentation} 
\label{sec:BrTS}
\noindent

The \benchmarkModel established the ``\fixedPerfRef'' ($\Pref = 0.726$).
To define the safety boundaries for iterative updates, we established a non-inferiority margin ($\tauval = 0.05$), resulting in a deterministic acceptance threshold of 0.676 ($=\Pref - \tauval$).

\begin{table}[ht]
    \small
\centering
\caption{\gls{dsc} Metric for Tumor Core Segmentation with Routing Trace}
\label{tab:dice_results}
\begin{tabularx}{\textwidth}{@{}
>{\centering\arraybackslash}m{.3cm}
>{\centering\arraybackslash}m{.7cm}
>{\centering\arraybackslash}m{1.5cm}
>{\centering\arraybackslash}m{1.5cm}
>{\centering\arraybackslash}m{3.5cm}
>{\centering\arraybackslash}m{1.8cm}
>{\centering\arraybackslash}m{.8cm}
>{\centering\arraybackslash}X
@{}}

\toprule
\textbf{Iter.} & \textbf{$\lvert \trainData \rvert$} & \textbf{$\Pcur$} & \textbf{$\PDcur$} & \textbf{Trigger} & \textbf{Routing Trace*} & \textbf{Deploy. \mbox{Decision}} & \textbf{\gls{pms} Signal} \\
\midrule
0 & 1150 & 0.726$\pm$0.313 & --- & \fixedPerfRef & --- & --- & ---\\
1 & 200 & 0.646$\pm$0.337 & 0.694$\pm$0.325 & safety violation ($P_{G, 1} < 0.676$)& $\rightarrow P1$ & REJ. & ---\\
2 & 400 & 0.673$\pm$0.337 & 0.748$\pm$0.289 & safety violation ($P_{G, 2} < 0.676$)& $\rightarrow P1$ & REJ. & ---\\
3 & 600 & 0.690$\pm$0.326 & 0.745$\pm$0.308 & minor drift ($\PgRel = 0.690$; $P_{D, 3} < P_{D, 2}$) & $P1\checkmark \rightarrow P4$ & APPR. & \alarm\\
4 & 800 & 0.700$\pm$0.321 & 0.758$\pm$0.284 & All satisfied ($\PgRel = 0.700$)& $P1\checkmark \rightarrow P4$ & APPR. & ---\\
5 & 1000 & 0.725$\pm$0.325 & 0.735$\pm$0.325 & minor drift ($\PgRel = 0.725$; $P_{D, 5} < P_{D, 4}$)& $P1\checkmark \rightarrow P4$ & APPR. & \alarm\\
6 & 1150 & 0.726$\pm$0.313 & 0.836$\pm$0.237 & All satisfied ($\PgRel = 0.726$) & $P1\checkmark \rightarrow P4$ & APPR. & ---\\
7 & 1350 & 0.742$\pm$0.308 & 0.892$\pm$0.200 & All satisfied ($\PgRel = 0.742$) & $P1\checkmark \rightarrow P4$ & APPR. & ---\\
8 & 1550 & 0.723$\pm$0.311 & 0.910$\pm$0.159 & $P_{G, 8} < \PgRel$ & $\rightarrow P1$ & REJ. & ---\\
9 & 1750 & 0.714$\pm$0.308 & 0.900$\pm$0.180 & $P_{G, 9} < \PgRel$; $P_{D, 9} < P_{D, 8}$ & $\rightarrow P1$ & REJ. & \alarm \\
10 & 1950 & 0.706$\pm$0.318 & 0.904$\pm$0.174 & $P_{G, 10} < \PgRel$ & $\rightarrow P1$ & REJ. & ---\\
11 & 2150 & 0.707$\pm$0.322 & 0.931$\pm$0.109 & $P_{G, 11} < \PgRel$ & $\rightarrow P1$ & REJ. & ---\\
12 & 2401 & 0.698$\pm$0.327 & --- & $P_{G, 12} < \PgRel$ & $\rightarrow P1$ & REJ. & ---\\
\bottomrule
\end{tabularx}

\vspace{0.3em}
\begin{flushleft}
    \footnotesize{Abbreviations: Iter. = iteration; APPR. = \approve; REJ. = \reject; Deploy. = Deployment.}\\
    \footnotesize{Notations: $\lvert \trainData \rvert$ = training set size; $\Pcur/\PDcur$ = \PgoldDrift.}\\
    \footnotesize{*Uses Table \ref{tab:decision_hierarchy} notation for priorities ($\checkmark=$ check passed; $\rightarrow$ = selected priority).}
\end{flushleft}
\end{table}

\begin{table*}[ht]
    \small
    \centering
    \caption{Mapping of Abstract \gls{aegis} Parameters to Brain Tumor Segmentation Parameters}
    \label{tab:bt_param_mapping}
    \begin{tabularx}{\textwidth}{@{} m{2.5cm} m{2.2cm} X m{3cm} @{}}
        \toprule 
        \multicolumn{1}{c}{\textbf{\gls{aegis} Parameter}} & \multicolumn{1}{c}{\textbf{Concrete Value}} & \multicolumn{1}{c}{\textbf{Clinical Justification}} & \multicolumn{1}{c}{\textbf{\gls{cdm} Decision}} \\
        \midrule
        $\Pref$ & \gls{dsc}: 0.726 & \fixedPerfRef. \benchmarkModelPerf (\gls{tc} region) & --- \\
        \midrule
        $\tauval$ & 0.05 & Defines $\Pfail$ boundary. Clinically acceptable margin below \fixedPerfRef & --- \\
        \midrule
        $\Pfail = \Pref - \tauval$ & 0.676 & Minimum acceptable \gls{tc} segmentation for treatment planning & --- \\
        \midrule
        $\cReg$ & $\Pcur < \Pfail$ & Performance falls below the minimum requirement for safe radiotherapy and surgical planning & \reject \\
        \midrule
        $\cNR$ & $\PgRel \leq \Pcur$ & Non-regression condition: must match or exceed \PgRelDef & \approve \\
        \midrule
        $\cReg$ & $\Pcur < \PgRel$ & Regression condition: must be less than \PgRelDef & \reject \\
        \midrule        
        $\cReg$ & $\PDcur < \PDprev$ & Early warning for data quality issues in incoming batch & \alarm \\
        \bottomrule
    \end{tabularx}
    \begin{flushleft}
        \footnotesize{Notations: $\Pref$ = \fixedPerfRef; $\tauval$ = \tolDef; $\Pfail$ = \PfailSaftyFloor; $\Pcur/\PDcur$ = \PgoldDrift; $\PgRel$ = \PgRelDefLong; $\cNR/\cReg$ = non-regression/regression condition.} \\
        \footnotesize{Note: \textit{\thrDisclaimer}}
    \end{flushleft}    
\end{table*} 

During the initial stages of the lifecycle, the segmentation performance for iterations 1 and 2 remained below this threshold, triggering an automatic \reject status.
In iteration 3, while the model's performance on the Golden set recovered to surpass the safety threshold, its performance on the drifting dataset declined relative to the previous step. 
This discrepancy produced an \approve + \alarm composite output: the deployment decision was \approve (Golden set above the \PfailSaftyFloor $\Pfail$), while the \gls{pms} \alarm was co-issued because the drifting set decline signaled a potential distribution shift in the newly assimilated batch warranting human adjudication and data forensic investigation.
The first successful model release occurred in iteration 4, where the system achieved an \approve status; here, the \gls{dsc} on the Golden set not only exceeded the threshold but showed improvement over all prior iterations, corroborated by a simultaneous performance gain on the drifting set. 
Following this logic, subsequent releases were authorized in iterations 6 and 7. 
In iteration 5, Golden set performance recovered above the \PgRelDef ($\PgRel$) but the drifting set declined, again producing \approve + \alarm, consistent with the same composite logic.
Notably, iteration 11 was classified as a \reject despite showing marginal gains over iteration 10 and significant improvement on the drifting set. 

This decision was governed by the \gls{pccp} logic: the Golden set performance remained approximately 3\% below the previously 
\PgRelDef (from iteration 7), failing the requirement for sustained or improved performance relative to the last \PgRelDef.
\section{Discussion} 
\noindent 

\subsection{Principal Findings}
\noindent
At its core, \gls{aegis} answers two questions independently at every iteration: 1) ``Can the new candidate model be deployed?'' and 2) ``Is the currently released model still safe in the field?'' The first is addressed by the deployment decision (P1--P4); the second by the parallel \gls{pms} \alarm signal (A1--A3).

This work presents \gls{aegis}, a governance \pccpConcept for continuously learning medical \gls{ai} systems. The modular architecture (\gls{darm}, \gls{mmm}, \gls{cdm}) provides manufacturers a structured approach to implementing \glspl{pccp} as defined by \gls{fda} guidance \cite{health_predetermined_2025}. 
Demonstration across the PhysioNet Sepsis Challenge dataset (20,336 patients across 11 iterative deployment cycles) established generalizability and practical feasibility.

The experimental results demonstrate five key \gls{aegis} capabilities (noting that these reflect governance mechanism behavior, not clinical validity):
\begin{enumerate}
    \item Maintenance of predetermined performance thresholds across normal operating conditions (iterations 0--5: sensitivity 0.723--0.753, specificity 0.926--0.933, \gls{roc-auc} 0.922--0.924)
    \item Co-issuance of the \gls{pms} \alarm signal at iteration 8, where the released model was assessed as at risk from major distributional shift (drift=1.000) even though the new candidate model passed all deployment checks (sensitivity 0.697). This demonstrates that the \gls{pms} monitoring channel operates independently of the deployment decision channel, and that distributional risk to the released model can be flagged without blocking deployment of an acceptable new model.
    \item Appropriate \reject decision when the \PfailSaftyFloor was violated (iteration 10: sensitivity 0.428 $<$ 0.65)
    \item The \glsposs{aegis} capacity to distinguish between data distribution changes (triggering the \gls{pms} \alarm signal for investigation) and actual safety violations (triggering \reject for the new candidate model)
    \item Exercise of all four deployment decision categories (\approve, \condApprvl, \clinRev, \reject) and both \alarm states (\alarm co-issued at iterations 8 and 10), including the \reject + \alarm composite at iteration 10---the critical safety escalation state where no deployable new model is available and the released model is simultaneously at risk. This composite state, which requires regulatory escalation beyond standard rollback, is unrepresentable in prior single-hierarchy frameworks.
\end{enumerate}

The brain tumor segmentation instantiation critically validates the \glsposs{aegis} capacity to govern high-dimensional spatial data, a primary challenge in adaptive radiology \gls{ai}. 
Unlike the tabular sepsis model, this validation exercised governance logic against complex distributional shifts inherent in multi-parametric \gls{mri}, such as changes in \gls{mri} acquisition parameters, as well as textural and morphological variances between pre-operative and post-treatment cohorts. 
An important regulatory finding occurred in iteration 3, where the system triggered an \alarm despite satisfactory Golden Set performance, correctly flagging a divergence in the incoming data stream that warranted investigation. 
Furthermore, the \reject decision in iteration 11, driven by a failure to match the previously \PgRelDef ($\PgRel$) rather than the absolute \PfailSaftyFloor, demonstrates the rigorous enforcement of the \gls{pccp} non-regression criteria.
This confirms that \gls{aegis} effectively prevents ``silent degradation'' in continuous learning loops, ensuring that new iterations maintain strict equivalence to the deployed \gls{sota} $\PgRel$ (\PgRelDefLong) rather than merely clearing minimum safety baselines.
Taken together, the two instantiations provide complementary coverage of the decision taxonomy: the sepsis case study exercises all four deployment categories plus both \alarm composite states through calibrated synthetic scenarios, while the brain tumor case study demonstrates that the same governance logic naturally produces \approve, \alarm, and \reject decisions from real distributional shifts across imaging cohorts.
Importantly, full taxonomy coverage is a property of the two instantiations \textit{jointly}, i.e., no single instantiation is required to exercise every category, because which categories are active is a deliberate configuration decision tied to the clinical context. The two case studies together confirm both that all categories are reachable through appropriate configuration and that selective activation is legitimate design practice.

The sepsis validation exercised all four deployment decision categories in the taxonomy (Table \ref{tab:decision_hierarchy}): \approve, \condApprvl, \clinRev, and \reject, plus \alarm composite states. The intermediate categories (\condApprvl via cross-site drift in iteration 6; \clinRev via regression from the \fixedPerfRef, $\Pref$, in iteration 7) demonstrate that the two-tier threshold architecture---hard \PfailSaftyFloor values combined with \fixedPerfRef quality checks---provides graduated governance responses between full approval and rejection. This graduated response pattern aligns with \gls{fda}'s emphasis on monitoring ``reasonably foreseeable risks'' in \gls{pccp} design.

\subsection{Comparison with Existing Approaches}
\noindent
Several governance frameworks address different layers. The \gls{abcds} framework from Duke Health provides institutional oversight through committee governance \cite{bedoya_framework_2022}. It emphasizes human decision-making at milestones but does not specify automated metrics. The \glsposs{fda} \gls{ai} Lifecycle concept offers phase-based guidance without implementation specifications \cite{health_blog_2025}. Babic et al.\ identified MAUDE gaps and recommended quarterly updates on drift and stability, but acknowledged undefined technical infrastructure \cite{babic_general_2025}.

\gls{aegis} provides what we term \textit{operational implementation}: concrete specifications (metrics, thresholds, decision logic, data protocols) translating high-level policies into executable procedures. Table \ref{tab:framework_comparison} summarizes differentiators.

\begin{table*}[t]
    \small
    \centering
    \caption{Comparison with Existing Approaches}
    \label{tab:framework_comparison}
    \begin{tabularx}{\textwidth}{@{} m{2.2cm} X X X X @{}}
        \toprule
        \multicolumn{1}{c}{\textbf{Dimension}} & \multicolumn{1}{c}{\textbf{\gls{abcds} (Duke)}} & \multicolumn{1}{c}{\textbf{\gls{fda} AILC}} & \multicolumn{1}{c}{\textbf{Babic et al.}} & \multicolumn{1}{c}{\textbf{\gls{aegis}}} \\
        \midrule
        Focus & Institutional governance & Conceptual lifecycle & Surveillance gaps & Operational implementation \\
        \midrule
        Contribution & Committee structure & Phase-based model & Policy recommendations & Metrics-driven logic \\
        \midrule
        Decision Mechanism & Committee review & Phase guidance & Quarterly updates & Threshold-based, five categories \\
        \midrule
        Monitoring & Periodic human review & Conceptual & Recommended & Continuous with drift detection \\
        \midrule
        Drift Detection & Not specified & Conceptual & Recommended & KS-test with threshold \\
        \midrule
        Regulatory Mapping & Implicit & General & \gls{fda} MAUDE & Explicit \gls{pccp}/\gls{mdr}/\gls{ai} Act \\
        \bottomrule
    \end{tabularx}
\end{table*}

The approaches are complementary. An organization might adopt \gls{abcds} for strategic oversight while deploying \gls{aegis} for operational monitoring.

Table \ref{tab:babic_mapping} maps Babic et al.\ recommendations to \gls{aegis} implementation.

\begin{table*}[t]
    \small
    \centering
    \caption{Mapping Babic et al.\ Recommendations to \gls{aegis}}
    \label{tab:babic_mapping}
    \begin{tabular}{m{3.9cm} m{8.5cm}}
        \toprule
        \multicolumn{1}{c}{\textbf{Recommendation}} & \multicolumn{1}{c}{\textbf{\gls{aegis} Implementation}} \\
        \midrule
        Quarterly deployment updates & \gls{darm} iterative cycle; Table \ref{tab:iteration_summary} shows 11 tracked iterations \\
        \midrule
        Concept drift detection & \gls{mmm} \gls{ks}-test (threshold $>$ 0.90) triggers \alarm \\
        \midrule
        Covariate shift monitoring & Golden dataset static; drifting dataset compared against reference \\
        \midrule
        Algorithmic stability & \fixedPerfRef comparison (1\% tolerance) \\
        \midrule
        Subgroup tracking & \gls{mmm} bias score across demographics \\
        \bottomrule
    \end{tabular}
\end{table*}

\subsection{Trustworthy \gls{ai} Dimensions}
\noindent
While primary governance decisions use clinical performance metrics (sensitivity, specificity, \gls{roc-auc}),  the \gls{aegis} architecture accommodates the full spectrum of Trustworthy \gls{ai} requirements increasingly mandated by regulatory bodies. \gls{aegis} implementation demonstrates this extensibility through concurrent monitoring of: (1) explainability metrics derived from \gls{shap} analysis, quantifying how concentrated predictive influence is among top features; (2) fairness metrics assessing performance parity across demographic subgroups; (3) calibration quality via Brier scores and the relationship between \gls{ppv}/\gls{npv} and clinical prevalence; and (4) statistical robustness indicators including Matthews Correlation Coefficient and Cohen's Kappa that are less sensitive to class imbalance than accuracy-based metrics.

For applications subject to \gls{eu} \gls{ai} Act transparency requirements (Article 13) or \gls{fda} guidance on clinical decision support transparency, manufacturers could formalize these dimensions as additional \gls{pccp} boundaries. For example, triggering \condApprvl when explainability scores drop below threshold (indicating the model's reasoning has shifted), or \alarm when fairness metrics reveal emerging disparities across protected groups. The \gls{npv} threshold ($\geq$0.90 in this demonstration) exemplifies clinically-grounded governance: for sepsis screening, high \gls{npv} ensures that patients receiving negative predictions can be confidently triaged to lower-acuity pathways. This multi-dimensional approach addresses regulatory expectations extending beyond traditional performance metrics.

\subsection{Limitations}
\noindent
\textbf{Scope Disclaimer:} The sepsis prediction and brain tumor segmentation applications serve exclusively as proof-of-concept demonstrations of governance \pccpConcept functionality. \textbf{No clinical evaluation or validation has been performed}, and the underlying predictive models are \textbf{not intended for clinical use}. 
Performance metrics reflect \glsposs{aegis} behavior under controlled conditions, not clinical efficacy or safety. Clinical deployment would require independent validation studies, \gls{irb} approval, and regulatory clearance commensurate with the system’s intended use and risk classification. Thresholds, weights, and decision boundaries are illustrative; production implementations must derive parameters through formal clinical risk assessment.

Several additional limitations warrant acknowledgment.

First, synthetic drift scenarios (iterations 6--10), while useful for demonstrating all four deployment decision categories and \alarm composite states, may not fully capture real-world complexity from demographic changes, equipment variations, or clinical practice evolution. The cross-site scenario (iteration 6) uses real Hospital B data from the PhysioNet Challenge, providing more realistic distributional shift than purely synthetic perturbations, but future validation should incorporate documented real-world drift events.

Second, \gls{aegis} relies on aggregate performance metrics that may not capture shifts in model uncertainty distribution. A model can maintain identical aggregate performance while dramatically shifting where it is uncertain. For ensemble-based systems, disagreement patterns between component models could provide additional governance signals; this extension represents a direction for future work.

Third, the current fairness assessment (bias score as sensitivity/specificity difference between demographic groups) is simplified. 
Production implementations should incorporate fairness evaluation across subgroups, consistent with the bias-mitigation and representativeness principles outlined in the \glsposs{fda} \gls{gmlp} guidance \cite{health_good_2025}.

Fourth, operational deployment considerations (computational resources, storage requirements for versioned datasets, human oversight burden) were not quantitatively evaluated. These factors significantly impact implementation feasibility.

Fifth, validation models do not represent \gls{sota} performance. This is intentional: demonstrating governance with conventional models establishes minimum functionality. 
Since \gls{aegis} interacts with models only through outputs and standard metrics, no architectural dependency prevents application to sophisticated models.

Sixth, aggregate \gls{kpi} monitoring may not detect adversarial data poisoning attacks that preserve overall performance while corrupting specific subpopulations or decision boundaries \cite{abtahi_data_2026}. Disagreement-based monitoring using temporal ensembles, comparing current models against historically validated versions, offers a promising detection mechanism \cite{abtahi_leveraging_2026}, though this approach requires validation in adversarial settings. The demonstrated detection delays when relying solely on performance metrics motivate development of complementary security signals.

Seventh, the brain tumor case study monitors a single performance metric (\gls{dsc} on \gls{tc}) without concurrent fairness, explainability, or composite \gls{mlcps} tracking, in contrast to the sepsis instantiation's comprehensive multi-dimensional monitoring. This asymmetry reflects the proof-of-concept scope: the brain tumor instantiation demonstrates the \glsposs{aegis} applicability to high-dimensional imaging data and its capacity to detect distribution shift through dual-dataset monitoring (Golden vs.\ Drifting). Future work should extend the brain tumor instantiation to include demographic subgroup analysis (e.g., tumor grade, patient age), explainability metrics appropriate for segmentation (e.g., saliency maps, uncertainty heatmaps), and a domain-appropriate \gls{mlcps} configuration to achieve monitoring parity with the sepsis case study.

\subsection{The \gls{mlcps} Metric}
\noindent
The \gls{mlcps} metric \cite{akshay_mlcps_2023} provides a principled approach to composite performance assessment by projecting multiple evaluation metrics onto a two-dimensional polar coordinate system. However, as noted by the original authors, the reliability of \gls{mlcps} depends on the quality of constituent metrics and the appropriateness of assigned weights. In clinical deployment, weight assignments should be validated through \thrSource rather than heuristic selection. The sensitivity-weighted configuration used in this demonstration (raw weights---Sensitivity: 1.5, \gls{roc-auc}: 1.3, Balanced Accuracy: 1.1, Specificity: 1.0---normalized internally by the official \gls{mlcps} package) reflects clinical priorities for sepsis detection; alternative applications may require substantially different weighting schemes derived through stakeholder consensus.

\subsection{Regulatory Implications}
\noindent
\gls{aegis} is structured to align with the \gls{fda}'s \gls{pccp} components as specified in the August 2025 final guidance \cite{health_marketing_2025, health_predetermined_2025}: Description of Modifications (\gls{darm} protocols, \gls{cdm} categories), Modification Protocol (\gls{mmm} procedures), and Impact Assessment (\fixedPerfRef comparison, audit trails). The four-category deployment taxonomy combined with the parallel \gls{pms} \alarm signal is designed to support documentation of predetermined governance pathways in 510(k) submissions, with the \alarm channel addressing separate \gls{mdr} \gls{pms} obligations. Whether a specific implementation satisfies submission requirements depends on device classification, intended use, and regulatory counsel.

For \gls{eu} manufacturers, predetermined changes documented in technical documentation are designed to support qualification for \gls{ai} Act exemptions from ``substantial modification'' requirements \cite{noauthor_mdcg_nodate}. Manufacturers should engage early with regulators to align parameters with expectations for specific device classifications.

\subsection{Generalizability}
\noindent
The choice of two maximally different clinical domains, structured tabular data for sepsis prediction versus high-dimensional imaging data for segmentation, was designed to stress-test \gls{aegis} generalizability. That the problem-specific instantiations from abstract \gls{darm}/\gls{mmm}/\gls{cdm} architectures handle both domains suggests applicability to other supervised learning tasks in healthcare, including natural language processing applications, signal processing (ECG, EEG), and multimodal fusion systems. 
Some key parameters for domain-specific adaptation or instantiation include threshold calibration (derived from clinical risk assessment) and \gls{mlcps} weight adjustment (reflecting clinical priorities), among others, as summarized in Table \ref{tab:instantiation_config}.
However, generalizability of the governance \pccpConcept does not imply clinical validity of any specific model deployed within it; each clinical application requires independent validation appropriate to its intended use.

\subsection{Adversarial Threats}
\noindent
The current \gls{aegis} implementation addresses natural drift, that is, distributional changes arising from demographic shifts, equipment upgrades, or evolving clinical practice, but does not explicitly address \textit{adversarial} threats. Data poisoning attacks, wherein malicious or corrupted samples are introduced into post-market data streams, represent a foreseeable risk that regulators increasingly expect manufacturers to anticipate \cite{abtahi_data_2026}. Critically, poisoning attacks can be designed to preserve aggregate performance metrics while degrading model behavior on targeted subpopulations or specific decision boundaries. This evasion of standard \gls{kpi} monitoring renders conventional \gls{pms} inadequate for security-critical applications.

We propose extending the \glsposs{aegis} monitoring layer to incorporate \textit{disagreement-based} signals adapted from ensemble learning approaches \cite{abtahi_leveraging_2026}. Rather than relying solely on aggregate metrics, the extended \gls{aegis} \pccpConcept would maintain previously validated model versions in a ``shadow'' or ``sentinel'' mode, continuously comparing their predictions against the current deployed model. Sudden increases in inter-version disagreement, particularly when localized to specific patient subgroups or feature regions, could serve as early indicators of data integrity compromise, thereby triggering \alarm or \reject decisions before poisoned data propagates into training sets. This temporal ensemble approach leverages the observation that poisoning attacks designed to evade aggregate metrics typically cannot simultaneously preserve agreement patterns across model generations.

Integration with the existing \gls{cdm} decision logic is straightforward: disagreement metrics would augment the current drift score as an additional input to the four-category deployment decision taxonomy and parallel \gls{pms} \alarm signal. However, the signal-to-action mapping requires careful calibration; disagreement spikes without corroborating adverse events might warrant enhanced monitoring rather than immediate rollback. The threshold configuration for disagreement-triggered governance decisions, like performance thresholds, should be derived through \thrSource and documented in the \gls{pccp} submission. Validation of this security-aware extension against realistic poisoning scenarios represents an important direction for future work.

\subsection{Future Directions}
\label{sec:future}
\noindent
Future work should address several practical extensions. Incorporation of uncertainty quantification methods would strengthen confidence estimation for release decisions by providing per-prediction reliability indicators alongside aggregate metrics. 
Extension to handle concept drift (changes in the relationship between features and outcomes) in addition to covariate drift would enhance robustness, as the current \pccpConcept primarily detects distributional changes in input features rather than shifts in outcome relationships \cite{m_s_survey_2023}. 

For security-aware governance, empirical characterization of reference disagreement rates between model versions would help distinguish normal model evolution from anomalous divergence patterns indicative of data integrity issues. Formal mapping of disagreement signals to \gls{cdm} decision categories, specifying when elevated disagreement should trigger \clinRev versus \alarm versus \reject, represents essential work for operationalizing the temporal ensemble approach.

Development of standardized benchmark datasets and drift scenarios specifically designed for \gls{pccp} framework evaluation would facilitate reproducible comparison of governance approaches across the research community.

\appendices

\section{Notation and Acronyms}
\label{app:notation}
For ease of reference, Table \ref{table:notation} summarizes the mathematical notations and acronyms used throughout this manuscript, respectively.
\begin{table}[h]
    \small
    \centering
    \caption{Mathematical Notation}
    \label{table:notation}
    \begin{tabular}{m{1.5cm} m{9.5cm}}   
        \toprule
        \multicolumn{1}{c}{\textbf{Symbol}} & \multicolumn{1}{c}{\textbf{Definition}} \\
        \midrule
        $\tauval$ & Tolerance: acceptable deviation from target value \\
        $\RaccG/\RaccD$ & Acceptable performance ranges for Golden/Drifting datasets \\
        $\Pref$ & \makefirstuc{\fixedPerfRef}: \fixedPerfRefLong. Set at first \approve; never updated. \\
        $\Pfail/\PPMS$ & \makefirstuc{\PfailPMS} \\        
        $\Pcur/\PDcur$ & \makefirstuc{\PgoldDrift} \\
        $\PgRel/\PdRel$ & \makefirstuc{\PgdRel} \\
        $\driftScore$ & \driftScoreDef \\
        $\driftScoreMinor/\driftScoreMajor$ & \driftScoreMinorMajor\\
        $\taiScore$ & Trustworthy AI composite score \\   
        $\taiScoreThr$ & Threshold for $\taiScore$ \\ 
        $\cSub{i}$ & General atomic condition $i$ in the \gls{cdm} Boolean function \\
        $\cNR/\cReg$ & Non-regression/regression conditions in the \gls{cdm} Boolean function \\
        $\dSub{}$ & \gls{cdm} decision output \\
        $\condFunc$ & Aggregated compound Boolean function over conditions \\
        $\CImarging$ & Non-inferiority margin for \gls{ci}-based equivalence testing \\
        $\trainData/\driftData$ & Training/Drifting dataset at iteration $k$ \\
        $\goldData$ & Golden (static held-out benchmark) dataset \\
        $K$ & Number of key clinical features for drift detection \\
        \bottomrule
    \end{tabular}
\end{table}

\setlength{\glsdescwidth}{0.9\linewidth}
\printnoidxglossary[type=acronym,title={Acronyms},style=long] 
\section*{Acknowledgments}
\noindent 
This study was supported by the Stockholm Medical Artificial Intelligence and Learning 
Environments (SMAILE) core facility at Karolinska Institutet and by \gls{eu} grant 240037 
from EIT Health as part of the Moodmon project, an \gls{ai}-powered digital monitoring 
platform for proactive, personalised care in affective disorders. The authors thank 
Małgorzata Sochacka, Tomasz Dominiak, and Zuzanna Czerwińska (Britenet Med) for sharing 
a compelling real-world use case that clearly illustrated the practical need for systematic 
model retraining in deployed clinical AI systems, which helped inspire the governance 
framework presented in this work.

\section*{Data Availability}
\noindent 
All data used in this study are publicly available through the cited sources.

\section*{Code Availability}
\noindent 
All code is available through the GitHub repository: \url{https://github.com/ki-smile/aegis} 

\section*{Author Contributions}
\noindent 
F. Afdideh: Methodology, Software, Validation, Formal analysis, Investigation, Data Curation, Writing -- Original Draft, Writing -- Review \& Editing, Visualization. 
M. Astaraki: Methodology, Software, Validation, Formal Analysis, Investigation, Data Curation, Writing -- Original Draft, Writing -- Review \& Editing, Visualization. 
F. Seoane: Methodology, Validation, Resources, Writing -- Review \& Editing, Supervision.
F. Abtahi: Conceptualization, Methodology, Software, Validation, Formal analysis, Investigation, Resources, Data Curation, Writing -- Original Draft, Writing -- Review \& Editing, Visualization, Supervision, Project administration, Funding acquisition.

\section*{Competing Interests}
\noindent 
The authors declare no competing interests.

\section*{Ethics Considerations}
\noindent 
This study used fully anonymized and public datasets. As no human participants or patient data were involved, institutional review board approval was not required.

\section*{Use of Artificial Intelligence Tools}
\noindent 
The authors used AI-based tools during manuscript preparation. Large language models were employed for English proofreading and readability improvement. The authors independently verified all cited sources. All scientific interpretations remain the sole responsibility of the authors.

\section*{Supplementary Information}
\noindent 
Supplementary information accompanies this paper, organized as three practical implementation resources:
\begin{itemize} 
    \item \textbf{Supplement S1 (Implementation Checklist):} Verification tool covering pre-implementation requirements, \gls{darm}/\gls{mmm}/\gls{cdm} configuration, regulatory documentation, and \gls{capa}/vigilance integration (Sections S1.1–-S1.6).
\item \textbf{Supplement S2 (Configuration Template):} Fillable template for documenting domain-specific instantiation parameters, including threshold specification with \isoRisk traceability, \gls{mlcps} weight configuration, \fixedPerfRef documentation, decision logic specification, regulatory pathway mapping, and risk acceptability matrix (Sections S2.1-–S2.8).
\item \textbf{Supplement S3 (Implementation Guide):} Step-by-step deployment guidance including \pccpConcept adoption workflow, threshold derivation from risk assessment, decision logic configuration with complete \gls{cdm} pseudo-code, Trustworthy \gls{ai} integration, regulatory submission guidance for \gls{fda} \gls{pccp} and \gls{eu} \gls{mdr}, post-deployment operations, and \gls{pms}/\gls{psur} reporting integration (Sections S3.1–-S3.7).
\end{itemize}
Appendices provide a completed example configuration for sepsis prediction (Appendix A), brain tumor segmentation (Appendix B), a decision category quick reference (Appendix C), and a \gls{capa} form template (Appendix D).
\bstctlcite{IEEEexample:BSTcontrol}
\bibliographystyle{IEEEtran/bibtex/IEEEtran}
\bibliography{IEEEtran/bibtex/IEEEabrv, bib/references}
\end{document}